\documentclass[11pt]{article}

\usepackage[preprint]{acl}

\usepackage{times}
\usepackage{latexsym}
\usepackage{amsmath}
\usepackage{cleveref}
\usepackage{float}
\usepackage{multirow}
\usepackage{makecell}

\RequirePackage{color}

\usepackage{todonotes}
\usepackage{soul}
\usepackage[T1]{fontenc}
\usepackage[utf8]{inputenc}

\usepackage{microtype}

\usepackage{inconsolata}

\usepackage{graphicx}
\usepackage{caption}
\usepackage{subcaption}
\usepackage{booktabs}
\usepackage{adjustbox}
\usepackage{multirow}
\usepackage{enumitem}
\usepackage{longtable}
\usepackage{tabularx}
\usepackage{xcolor}
\usepackage{array}
\usepackage{tabularx}
\usepackage{longtable}

\definecolor{TopShade}{RGB}{210,220,255}     % light bluish
\definecolor{BaseShade}{RGB}{255,215,215}    % light reddish
\definecolor{BlockBorder}{gray}{0.25}

\newcolumntype{L}{>{\raggedright\arraybackslash}X}

\title{SemEval-2026 Task 3: Dimensional Aspect-Based Sentiment Analysis (DimABSA)}

\author{
\textbf{Liang-Chih Yu}$^{1,*}$,
\textbf{Jonas Becker}$^{2,*}$,
\textbf{Shamsuddeen Hassan Muhammad}$^{3}$,
\textbf{Idris Abdulmumin}$^{4}$, \\
\textbf{Lung-Hao Lee}$^{5,*}$,
\textbf{Ying-Lung Lin}$^{6}$,
\textbf{Jin Wang}$^{7}$,
\textbf{Jan Philip Wahle}$^{2}$,
\textbf{Terry Ruas}$^{2}$, \\
\textbf{Natalia Loukachevitch}$^{8}$,
\textbf{Alexander Panchenko}$^{9,10}$,
\textbf{Ilseyar Alimova}$^{9}$,
\textbf{Lilian Wanzare}$^{11}$, \\
\textbf{Nelson Odhiambo}$^{11}$, 
\textbf{Bela Gipp}$^{2}$,
\textbf{Kai-Wei Chang}$^{12}$, 
\textbf{and Saif M. Mohammad}$^{13}$ \vspace{+0.5em}
\\
$^{1}$Yuan Ze University,
$^{2}$University of G\"ottingen,
$^{3}$Imperial College London,
$^{4}$University of Pretoria, \\
$^{5}$National Yang Ming Chiao Tung University,
$^{6}$Central Police University, \\
$^{7}$Yunnan University,
$^{8}$Lomonosov Moscow State University,
$^{9}$Skoltech,
$^{10}$AIRI, \\
$^{11}$Maseno University,
$^{12}$UCLA, 
$^{13}$National Research Council Canada \vspace{+0.5em}
\\
\textbf{\small *Equal contribution} \\
{\small \textbf{Contact:} \href{mailto:lcyu@saturn.yzu.edu.tw}{lcyu@saturn.yzu.edu.tw}, \href{mailto:jonas.becker@uni-goettingen.de}{jonas.becker@uni-goettingen.de}}
}
\begin{document}
\maketitle
\begin{abstract}
We present the SemEval-2026 shared task on Dimensional Aspect-Based Sentiment Analysis (DimABSA), which improves traditional ABSA by modeling sentiment along valence–arousal (VA) dimensions rather than using categorical polarity labels. To extend ABSA beyond consumer reviews to public-issue discourse (e.g., political, energy, and climate issues), we introduce an additional task, Dimensional Stance Analysis (DimStance), which treats stance targets as aspects and reformulates stance detection as regression in the VA space. The task consists of two tracks: Track A (DimABSA) and Track B (DimStance). Track A includes three subtasks: (1) dimensional aspect sentiment regression, (2) dimensional aspect sentiment triplet extraction, and (3) dimensional aspect sentiment quadruplet extraction, while Track B includes only the regression subtask for stance targets. We also introduce a continuous F1 (cF1) metric to jointly evaluate structured extraction and VA regression.

The task attracted more than 400 participants, resulting in 112 final submissions and 42 system description papers. We report baseline results, discuss top-performing systems, and analyze key design choices to provide insights into dimensional sentiment analysis at the aspect and stance-target levels. All resources are available on our GitHub repository\footnote{\url{https://github.com/DimABSA/DimABSA2026}.}
\end{abstract}

%The task comprises two tracks: (A) DimABSA and (B) Dimensional Stance Analysis (DimStance). Track A covers six languages -- \textit{English, Japanese, Russian, Tatar, Ukrainian, and Chinese} -- across four domains: restaurant, laptop, hotel, and finance. Track B includes five languages -- \textit{English, German, Chinese, Nigerian Pidgin, and Swahili} -- spanning environmental protection and politics. In total, the datasets contain 76,958 aspect instances across 42,590 sentences for DimABSA and 11,746 stance targets across 7,365 texts for DimStance. Participants competed in three subtasks: (1) dimensional aspect sentiment regression, (2) dimensional aspect–opinion triplet extraction, and (3) dimensional aspect quadruplet extraction, with Track B focusing on the regression setting. 

\section{Introduction}
Aspect-Based Sentiment Analysis (ABSA) is a widely used technique for analyzing opinions and sentiments at the aspect level. It is formulated as the extraction of sentiment elements, including aspect terms, aspect categories, opinion terms, and sentiment polarity, individually or jointly. For example, given the sentence \textit{The food was excellent.}, an ABSA system is expected to extract the aspect term \textit{food}, the opinion term \textit{excellent}, assign the aspect category {\small \texttt{FOOD\#QUALITY}} from a predefined set, and predict {\small \texttt{Positive}} sentiment polarity. Following the success of prior SemEval tasks \cite{pontiki-etal-2014-semeval,pontiki-etal-2015-semeval,pontiki2016semeval}, ABSA has attracted substantial attention, providing deeper insights into user opinions across various applications \cite{d2022knowmis, Zhang2023, hua2024systematic}.

\begin{figure}[t]
    \centering
    \includegraphics[width=0.9\linewidth, trim=10 10 10 10, clip]{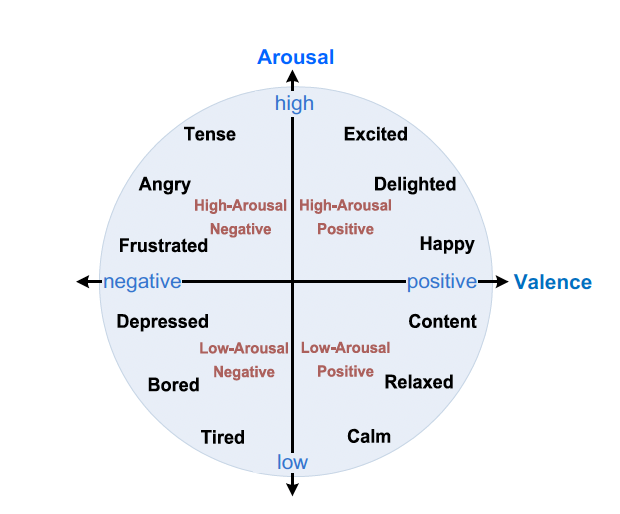}
    \caption{Valence--Arousal (VA) space.}
    \label{fig:VAspace}
\end{figure}

However, current ABSA research adopts a coarse-grained, categorical sentiment representation (e.g., positive, negative, and neutral). This approach contrasts with long-established theories in psychology and affective science \cite{russell1980circumplex, russell2003core}, where sentiment is represented along fine-grained, real-valued dimensions of \textbf{valence} (from negative to positive) and \textbf{arousal} (from sluggish to excited), as illustrated in \Cref{fig:VAspace}. This valence--arousal (VA) representation has motivated research on dimensional sentiment analysis \cite{Yu2016, buechel-hahn-2017-emobank, Mohammad2018semeval, Lee2022, Lee2024, mohammad2025nrcvad}, enabling more nuanced distinctions in emotional expression and supporting broader applications.

To move beyond categorical sentiment labels, we introduce a SemEval shared task that integrates the dimensional VA representation into the traditional ABSA framework. We refer to this task as Dimensional ABSA (DimABSA). To this end, we construct multilingual, multi-domain datasets by annotating traditional ABSA elements (aspect terms, aspect categories, and opinion terms) together with continuous VA scores \cite{lee2026dimabsabuildingmultilingualmultidomain}. 

Furthermore, stance detection and ABSA are conceptually related, as stance targets can be treated as aspects. Building on this connection, we introduce an additional task, Dimensional Stance Analysis (DimStance), which requires systems to predict VA scores for given targets. For this task, we annotate stance targets with VA scores to construct multilingual, multi-domain datasets \cite{becker2026dimstancemultilingualdatasetsdimensional}. The DimStance formulation not only extends ABSA beyond consumer reviews to public-issue discourse (e.g., political, energy, and climate issues) but also generalizes stance analysis from categorical labels to the VA representation. 

We organize the SemEval task into two tracks: Track A (DimABSA) and Track B (DimStance). We further design three subtasks that combine VA scores with different ABSA elements: (1) Dimensional Aspect Sentiment Regression (DimASR), predicting VA scores for each aspect in a sentence; (2) Dimensional Aspect Sentiment Triplet Extraction (DimASTE), jointly extracting aspect and opinion terms and predicting their associated VA scores; and (3) Dimensional Aspect Sentiment Quadruplet Extraction (DimASQP), extending DimASTE by additionally predicting aspect categories. Track A includes all three subtasks, while Track B includes only Subtask 1 (DimASR). 

Our task attracted over 400 participants, resulting in 112 final submissions from 44 teams and 42 system description papers. Track A (DimABSA) was the most popular, with over 300 participants and 84 final submissions, while Track B (DimStance) attracted over 100 participants with 28 final submissions. Notably, most teams participated in multilingual and multidomain settings, covering an average of \textasciitilde 4.5 languages and \textasciitilde 3.4 domains.

Analysis of participating systems reveals that most approaches leverage pretrained transformers or large language models (LLMs). These models are typically trained with supervised fine-tuning and enhanced with various training and prompting strategies. Evaluation results show that dimensional sentiment analysis at the aspect and stance-target levels remains challenging, especially for low-resource languages.

%Across the two tracks, the shared task attracted broad participation and coverage. Track A saw participation from 40 teams, while Track B included 13 teams, with 10 teams contributing to both tracks. Track A spanned six languages (English, Japanese, Russian, Tatar, Ukrainian, and Chinese) and four domains (Finance, Hotel, Laptop, and Restaurant), whereas Track B focused on five languages (German, English, Nigerian Pidgin, Swahili, and Chinese) across two domains (Environment and Politics), highlighting the multilingual and multidomain nature of the shared task.

\section{Related Work}
\paragraph{Categorical ABSA.} Most existing ABSA datasets are English-centric and primarily focus on customer review applications \cite{chebolu-etal-2023-review}. SemEval-2014 \cite{pontiki-etal-2014-semeval} introduced the first ABSA shared task for English restaurant and laptop reviews, followed by extensions to additional subtasks and languages \cite{pontiki-etal-2015-semeval, pontiki2016semeval}. 
Subsequent datasets further enriched the annotation schema, introducing triplets of aspect, opinion, and polarity \cite{Xu2020, Peng2020}, and quadruples by adding an aspect category \cite{Zhang2021, Cai2021}. The domain coverage has also been broadened to areas such as finance \cite{Kubo2018}, COVID-19 \cite{aygun2022aspect, 2025_Hou}, and education \cite{Hua2025EduRABSA}. Moreover, M-ABSA \cite{Wu2025} extended this line of work to the multilingual setting by constructing a parallel benchmark through automatic translation.

\paragraph{Categorical Stance.} Prior work on stance detection has expanded primarily along three axes: language coverage, scale, and domain specificity. Early benchmarks focused on English Twitter data, such as the SemEval stance dataset \cite{MohammadKSZ16a}). Multilingual extensions followed, including X-Stance \citep{VamvasS20b} for German, French, and Italian, and the Catalonia Independence Corpus (CIC) for Catalan and Spanish \citep{ZotovaANR20}. Large-scale English resources, such as P-Stance \citep{LiSSN21} and COVID-19-Stance \cite{GlandtKLC21}, further increased dataset size and target diversity. Recent work has extended stance detection to zero-shot \cite{allaway-mckeown-2020-zero, zhao-etal-2023-c, zhao-caragea-2024-ez}, multimodal \cite{zhou-etal-2025-media-frames-stance, zhang-etal-2025-mad}, and conversational \cite{ding-etal-2025-zero, marreddy-etal-2025-usdc} settings.

\paragraph{Dimensional Sentiment Analysis.} Previous studies have developed resources with single- or combined-dimensional representations across lexical, phrasal, and sentential granularities. 
Sentiment lexicons assign affective scores to individual words, such as SentiWordNet \cite{Baccianella2010}, SO-CAL \cite{Taboada2011}, \mbox{SentiStrength} \cite{Thelwall2012}, and NRC-VAD \cite{mohammad2018obtaining, mohammad2025nrcvad}. 
Phrase-level datasets formulate sentiment composition through modifiers, including SemEval-2015 Task 10 \cite{Rosenthal2015} and SemEval-2016 Task 7 \cite{Kiritchenko2016}. 
At the sentence level, affective scores are provided for texts of varying lengths \cite{preotiuc-pietro-etal-2016-modelling, Buechel2017, mohammad-bravo-marquez-2017-emotion, Mohammad2018semeval, muhammad-etal-2025-brighter}. The Stanford Sentiment Treebank \cite{Socher2013} and Chinese EmoBank \cite{Lee2022} provide cross-granularity resources, bridging phrase- and sentence-level representations and covering all three granularities.

\section{Task Description}
\subsection{Track A: Dimensional Aspect-Based Sentiment Analysis (DimABSA)}
This track involves three traditional ABSA elements and VA scores, described as follows.

\noindent\textbf{Aspect Term (A)}: a word or phrase indicating an opinion target, such as \textit{service}, \textit{screen}, \textit{profit}. 

\noindent\textbf{Aspect Category (C)}: a predefined Entity\#Attribute label associated with an aspect term (e.g., {\small \texttt{FOOD\#QUALITY}}, {\small \texttt{SERVICE\#GENERAL}}) \cite{pontiki-etal-2015-semeval, pontiki2016semeval}. The full list of aspect categories is presented in Appendix A. 

\noindent\textbf{Opinion Term (O)}: a sentiment-bearing word or phrase associated with a specific aspect term. The opinion term includes sentiment modifiers to support fine-grained sentiment representation (e.g., \textit{very good}, \textit{extremely bad}, \textit{a little slow}).

\noindent\textbf{Valence-Arousal (VA)}: a pair of real-valued scores, each ranging from 1 to 9, where 1 denotes extreme negative valence or low arousal, 9 denotes extreme positive valence or high arousal, and 5 denotes neutral valence or medium arousal.

Based on these elements, we define three subtasks that adapt traditional ABSA formulations to the dimensional sentiment paradigm. We present in-/output formats and an example in Appendix \ref{sec:trackA-subtask-example}.

\begin{itemize}[noitemsep,leftmargin=*]
\item \textbf{Subtask 1 - Dimensional Aspect Sentiment Regression (DimASR):} Given a sentence and one or more aspects, predict VA scores for each aspect. This task generalizes traditional Aspect Sentiment Classification (ASC) \cite{pontiki-etal-2014-semeval, pontiki-etal-2015-semeval, pontiki2016semeval} to VA regression.

\item \textbf{Subtask 2 - Dimensional Aspect Sentiment Triplet Extraction (DimASTE):} Given a sentence, extract all \texttt{(A, O, VA)} triplets. This task jointly extracts aspect and opinion terms and predicts their associated VA scores, extending traditional ASTE \cite{Peng2020} by incorporating VA regression.

\item \textbf{Subtask 3 - Dimensional Aspect Sentiment Quadruplet Prediction (DimASQP):} Given a sentence, extract all \texttt{(A, C, O, VA)} quadruplets. Compared to DimASTE, this task additionally incorporates aspect category classification, extending traditional ASQP \cite{Cai2021, Zhang2021} to include VA regression.

\end{itemize}

\subsection{Track B: Dimensional Stance Analysis (DimStance)}
Given an utterance or post and a target entity, stance detection is formulated as determining whether the speaker is in favor of the target, against the target, or neither inference is likely \cite{mohammad2017stance}. This track reformulates stance detection by treating stance targets as aspects and generalizes categorical stance classification to VA regression. We adopt the formulation of Track A (DimASR), where the input is a text and one or more targets, and the task is to predict VA scores for each target.  

\section{Datasets}
We construct multilingual, multi-domain datasets for both Track A (DimABSA) and Track B (DimStance), as shown in \Cref{tab:merged_dimabsa_dimstance}.
Detailed descriptions of the data sources, annotation process, and annotation agreement are provided in our dataset papers \cite{lee2026dimabsabuildingmultilingualmultidomain, becker2026dimstancemultilingualdatasetsdimensional}. Key information is summarized below.

% Preamble (add once)
% \usepackage{booktabs}
% \usepackage{adjustbox}
% \usepackage{multirow}
% \usepackage{graphicx}

\begin{table*}[t]
\centering
\small
\setlength{\tabcolsep}{4pt}
\begin{adjustbox}{max width=\linewidth}
\begin{tabular}{c l l l | c c c c}
\toprule
\multirow{2}{*}[-0.3em]{\textbf{Track}} &
\multirow{2}{*}[-0.3em]{\textbf{Dataset}} &
\multirow{2}{*}[-0.3em]{\textbf{Source(s)}} &
\multirow{2}{*}[-0.3em]{\textbf{Subtask}} &
\multicolumn{4}{c}{\textbf{Number of texts / instances}} \\
\cmidrule(lr){5-8}
 &  &  &  &
\textbf{Train} & \textbf{Dev} & \textbf{Test} & \textbf{Total} \\
\midrule

% ======================
% Track A: DimABSA
% ======================
\multirow{18}{*}{\rotatebox{90}{\textbf{Track A}}}

& \multirow{2}{*}{eng-rest}
& \multirow{2}{*}{ACOS; Yelp Open Dataset}
& ST1
& \multirow{2}{*}{2284 / 3659}
& 200 / 340 & 1000 / 1504 & 3484 / 5503 \\
& & & ST2--3
& & 200 / 408 & 1000 / 2129 & 3484 / 6196 \\ \addlinespace[2pt]

& \multirow{2}{*}{eng-lap}
& \multirow{2}{*}{ACOS; Amazon Reviews 2023}
& ST1
& \multirow{2}{*}{4076 / 5773}
& 200 / 275 & 1000 / 1421 & 5276 / 7469 \\
& & & ST2--3
& & 200 / 317 & 1000 / 1975 & 5279 / 8065 \\ \addlinespace[2pt]

& \multirow{2}{*}{jpn-hot}
& \multirow{2}{*}{Rakuten Travel}
& ST1
& \multirow{2}{*}{1600 / 2846}
& 200 / 284 & 800 / 1092 & 2600 / 4222 \\
& & & ST2--3
& & 200 / 364 & 800 / 1443 & 2600 / 4653 \\ \addlinespace[2pt]

& jpn-fin
& chABSA; EDINET
& ST1
& 1024 / 1672 & 200 / 319 & 800 / 1302 & 2024 / 3293 \\ \addlinespace[2pt]

& \multirow{2}{*}{rus-rest}
& \multirow{2}{*}{SemEval\textquoteright16 Task 5 (Restaurant)}
& ST1
& \multirow{2}{*}{1240 / 2487}
& 56 / 81 & 1072 / 1637 & 2368 / 4205 \\
& & & ST2--3
& & 48 / 102 & 630 / 1310 & 678 / 3899 \\ \addlinespace[2pt]

& \multirow{2}{*}{tat-rest}
& \multirow{2}{*}{SemEval\textquoteright16 (MT)}
& ST1
& \multirow{2}{*}{1240 / 2487}
& 56 / 81 & 1072 / 1637 & 2368 / 4205 \\
& & & ST2--3
& & 48 / 102 & 630 / 1310 & 678 / 3899 \\ \addlinespace[2pt]

& \multirow{2}{*}{ukr-rest}
& \multirow{2}{*}{SemEval\textquoteright16 (MT)}
& ST1
& \multirow{2}{*}{1240 / 2487}
& 56 / 81 & 1072 / 1637 & 2368 / 4205 \\
& & & ST2--3
& & 48 / 102 & 630 / 1310 & 678 / 3899 \\ \addlinespace[2pt]

& \multirow{2}{*}{zho-rest}
& \multirow{2}{*}{SIGHAN-2024; Google Reviews; PTT}
& ST1
& \multirow{2}{*}{6050 / 8523}
& 225 / 416 & 1000 / 1929 & 7275 / 10868 \\
& & & ST2--3
& & 300 / 761 & 1000 / 2861 & 7350 / 12145 \\ \addlinespace[2pt]

& \multirow{2}{*}{zho-lap}
& \multirow{2}{*}{Mobile01}
& ST1
& \multirow{2}{*}{3490 / 6502}
& 261 / 431 & 1000 / 2662 & 4751 / 9595 \\
& & & ST2--3
& & 300 / 551 & 1000 / 2798 & 4790 / 9851 \\ \addlinespace[2pt]

& zho-fin
& MOPS
& ST1
& 1000 / 2633 & 200 / 563 & 842 / 2354 & 2042 / 5550 \\

\midrule

% ======================
% Track B: DimStance
% ======================
\multirow{5}{*}{\rotatebox{90}{\textbf{Track B}}}

& eng-env & EZ-STANCE; Reddit & ST1
& 922 / 2059  & 200 / 339 & 1020 / 1813 & 2142 / 4211 \\
& deu-pol & Wahl-O-Mat Archive & ST1
& 683 / 1335  & 34 / 75   & 263 / 438   & 980 / 1848 \\
& zho-env & Threads Platform & ST1
& 683 / 1091  & 34 / 49   & 263 / 898   & 980 / 2038 \\
& pcm-pol & X Platform & ST1
& 1049 / 1118 & 119 / 122 & 331 / 343  & 1499 / 1583 \\
& swa-pol & X Platform & ST1
& 1375 / 1622 & 123 / 145 & 266 / 299  & 1764 / 2066 \\

\bottomrule
\end{tabular}
\end{adjustbox}
\caption{\textbf{Dataset statistics for Track A (DimABSA) and Track B (DimStance).}
For each dataset (language–domain), we report the source(s), subtask type (ST1 vs.\ ST2–3), and the number of texts and instances in the train/dev/test splits, using the format \texttt{\#texts/\#instances}. There can be multiple instances per text.}
\label{tab:merged_dimabsa_dimstance}
\end{table*}

Track A covers six languages: English (\texttt{eng}), Japanese (\texttt{jpn}), Russian (\texttt{rus}), Tatar (\texttt{tar}), Ukrainian (\texttt{ukr}), and Chinese (\texttt{zho}). These datasets span four domains: restaurant (\texttt{rest}), laptop (\texttt{lap}), hotel (\texttt{hot}), and finance (\texttt{fin}). The finance datasets are used exclusively for Subtask 1, while the other domains support all three subtasks. In total, Track A provides 76,958 aspect instances (aspect pairs, triplets, and quadruplets) across 42,590 sentences.

Track B comprises five language-specific datasets:  English (\texttt{eng}), German (\texttt{deu}), Chinese (\texttt{zho}), Nigerian Pidgin (\texttt{pcm}), and Swahili (\texttt{swa}). These datasets support two domains: environmental protection (\texttt{env}) and politics (\texttt{pol}). They are used exclusively for Subtask 1. In total, Track B contains 11,746 stance targets across 7,365 texts. 

\subsection{Data Collection}

\subsubsection{Track A: DimABSA}
We collect data from multiple sources, including existing labeled ABSA datasets and newly curated unlabeled data. The existing labeled datasets are used solely for training, while the newly curated data are annotated and split into training, development, and test sets. The data sources for each language are described below.

\noindent\textbf{English.} We use the training split of the ACOS dataset \cite{Cai2021}, manually annotating the restaurant and laptop quadruplets with VA scores to replace the sentiment polarity labels. 
For the development and test sets, we collect restaurant reviews from Yelp Open Dataset\footnote{\href{https://business.yelp.com/data/resources/open-dataset}{https://business.yelp.com/data/resources/open-dataset}} and laptop reviews from Amazon Reviews 2023 \cite{hou2024bridging}.  

\noindent\textbf{Japanese.} For the finance domain, the training set is sampled from the chABSA dataset.\footnote{\href{https://github.com/chakki-works/chABSA-dataset}{https://github.com/chakki-works/chABSA-dataset}} 
We manually annotate VA scores for each aspect in these samples, replacing the original sentiment polarity labels. The development and test sets are collected from the same EDINET\footnote{\href{https://disclosure2.edinet-fsa.go.jp}{https://disclosure2.edinet-fsa.go.jp}} sources as chABSA, with samples involving the same companies removed to avoid overlap. For the hotel domain, we crawl reviews from Rakuten Travel.\footnote{\href{https://travel.rakuten.co.jp}{https://travel.rakuten.co.jp}}

\noindent\textbf{Russian.} The SemEval-2016 restaurant review dataset \cite{pontiki2016semeval} serves as the data source. The labeled portion contains annotated aspects, their categories, and sentiment polarity. For the remaining instances, opinion terms and VA values are annotated. The unlabeled portion of reviews is used for the development and test sets. 

\noindent\textbf{Tatar.} We automatically translate the Russian dataset into Tatar using Yandex Translate. The translations are then reviewed by a native speaker, manually correcting 45.5\% of instances.

\noindent\textbf{Ukrainian.} Similar to Tatar, we translated the Russian dataset into Ukrainian. 35.6\% of instances were manually corrected by native speakers.

\noindent\textbf{Chinese.} For the restaurant domain, we use the SIGHAN-2024 dataset \cite{Lee2024} for training, and construct the development and test sets from Google Reviews\footnote{\href{https://customerreviews.google.com}{https://customerreviews.google.com}} and the PTT platform\footnote{\href{https://www.pttweb.cc}{https://www.pttweb.cc}}. For the laptop domain, we crawl reviews from Mobile01\footnote{\href{https://www.mobile01.com/category.php?id=2}{https://www.mobile01.com/category.php?id=2}}. For the finance domain, we collect annual reports of Taiwanese companies from MOPS\footnote{\href{https://emops.twse.com.tw/server-java/t58query}{https://emops.twse.com.tw/server-java/t58query}}.

\subsubsection{Track B: DimStance}
\noindent\textbf{English.} Data for the training split is collected from the environmental protection domain of EZ-STANCE \citep{zhao-caragea-2024-ez}. The dev and test splits are obtained from Reddit texts\footnote{\url{https://reddit.com/}. Version: 2025-07-01}, using the same keywords as in EZ-STANCE.

\noindent\textbf{German.} Sampled from Wahl-O-Mat Archive, provided by the Federal Agency for Civic Education of Germany\footnote{\url{https://www.bpb.de/themen/wahl-o-mat/556865/datensaetze-des-wahl-o-mat/}. Version: 2026-03-25.}. The data contains responses by political parties to political statements.

\noindent\textbf{Chinese.} Collected messages from the Threads platform\footnote{\url{https://www.threads.com/}. Version: 2025-10-15}. Crawling is performed using a predefined set of Chinese query keywords about environmental protection. 

\noindent\textbf{Nigerian Pidgin.} Sampled posts and comments from the X platform\footnote{\url{https://x.com/}. Version: 2023-12-31}. The discussions concern Nigerian elections, i.e., politics, ranging from January 1st to March 8th, 2023.

\noindent\textbf{Swahili.} Combined data from Afrisenti \citep{muhammad-etal-2023-afrisenti}, HateSpeech\_Kenya \citep{Ombui}, and Politikweli \citep{Amol_PolitikWeli}, covering political tweets from the X platform.

\subsection{Annotation Process}
The annotated elements vary across datasets depending on the subtask configuration. We annotate \texttt{(A, VA)} pairs for datasets exclusive to \mbox{Subtask 1} (DimASR), specifically the finance datasets in Track A and all datasets in Track B. For datasets that support all subtasks, we annotate full \texttt{(A, C, O, VA)} quadruplets. This design facilitates a shared training set across subtasks, as indicated by the dataset splits in \Cref{tab:merged_dimabsa_dimstance}. However, we do not use a shared development/test set for all subtasks, as Subtask 1 assumes aspect terms are provided as input. Instead, we create a dedicated development/test set for Subtask 1, and a shared set for Subtask 2 (DimASTE) and Subtask 3 (DimASQP).

The annotation process is conducted in two phases: we first extract categorical tuples from sentences, identifying the element \texttt{A} for each \texttt{(A, VA)} pair and the triplet \texttt{(A, C, O)} for each quadruplet, followed by the assignment of VA scores.
For Track A, two annotators independently extract tuples from each sentence, with a third adjudicator resolving disagreements. For Track B, we use LLMs to extract candidate stance targets, which are then validated by five annotators through majority voting. For VA rating, both tracks rely on five annotators, and the final VA score is computed by averaging the ratings. 

\subsection{Annotation Quality}
We evaluate the agreement at the tuple level using the F1 score, following prior work~\cite{Chebolu2024, Wu2025}. The F1 score is computed between two annotators by treating one as the prediction and the other as the gold standard. To assess VA agreement, we use Root Mean Square Error (RMSE) separately for valence and arousal. RMSE is calculated by comparing each annotator's rating against the mean of all five annotators. The final agreement score is the average RMSE across all five annotators. 

\section{Evaluation}
\subsection{Metrics}
\paragraph{Subtask 1: RMSE.}
DimASR is formulated as a regression task, and its performance is evaluated by measuring prediction error in the VA space using RMSE, defined as
\vspace{-0.25cm}
\begin{equation}
\scriptsize
\mathrm{RMSE}_\mathrm{{VA}} =
\sqrt{
\frac{1}{N}
\sum_{i=1}^{N}
\left(V_{p}^{(i)}-V_{g}^{(i)}\right)^2+
\left(A_{p}^{(i)}-A_{g}^{(i)}\right)^2
}
\end{equation}

\noindent  where $N$ is the total number of instances; $V_p^{(i)}$ and $A_p^{(i)}$ denote the predicted valence and arousal values for an instance; and $V_g^{(i)}$ and $A_g^{(i)}$ denote the corresponding gold values.

\paragraph{Subtask 2 \& 3: Continuous F1 (cF1).}
DimASTE and DimASQP are hybrid tasks that require both categorical prediction and VA regression. Therefore, the standard F1 score, widely used in ABSA, is insufficient to jointly assess these components. To address this limitation, we propose the \textit{continuous F1 (cF1)} metric, which incorporates VA prediction error into the F1 formulation. 

Following the standard F1, a predicted tuple is counted as a true positive (TP) only if all its categorical elements exactly match the gold annotation. This categorical TP is then extended as a \textit{continuous true positive (cTP)} by incorporating the VA prediction error. Formally, let $P$ denote the set of predicted triplets \texttt{(A, O, VA)} or quadruplets \mbox{\texttt{(A, C, O, VA)}}. For any prediction $t \in P$, its cTP is defined as
\begin{equation}
\small
\mathrm{cTP}^{(t)} =
\begin{cases} 
1 - \mathrm{dist}(\mathrm{VA}_p^{(t)}, \mathrm{VA}_g^{(t)}), & t \in P_\mathrm{cat} \\
0, & \text{otherwise}
\end{cases}
\end{equation}

\noindent where $P_{cat} \subseteq P$ denotes the set of predictions in which all categorical elements, \texttt{(A, O)} for a triplet or \texttt{(A, C, O)} for a quadruplet, exactly match the gold annotation for the same sentence. The distance function is defined as
%Each categorically correct prediction \textbf{$t \in P_{cat}$} is assigned an initial TP score of 1, which is then reduced based on its VA error distance. 
\begin{equation}
\small
\mathrm{dist}(\mathrm{VA}_{p}, \mathrm{VA}_{g}) =
\frac{
\sqrt{(V_p - V_g)^2 + (A_p - A_g)^2}
}{
\mathrm{D}_{\max}
}
\end{equation}

\noindent where $\operatorname{dist}(\cdot)$ denotes the normalized Euclidean distance between the predicted $\mathrm{VA}_p = (V_p, A_p)$ and gold $\mathrm{VA}_{g}=(V_g,A_g)$ in the VA space, and $\mathrm{D}_{\max} = \sqrt{8^2+8^2} = \sqrt{128}$ is the maximum possible Euclidean distance in the VA space on the [1, 9] scale, ensuring that $\operatorname{dist} \in [0, 1]$.

Building on per-prediction $\mathrm{cTP}^{(t)}$, \textit{cRecall} and \textit{cPrecision} are defined as the total cTP divided by the number of gold and predicted triplets/quadruplets, respectively. The cF1 is computed as their harmonic mean. An illustrative example is given in Appendix~\ref{sec:example_calculating_cF1}. An official evaluation script is available on the task GitHub repository.  

\subsection{Baselines}
We provide two baseline systems for each track. 
\paragraph{Track A.} We employ the closed-source LLM Kimi K2 Thinking \cite{MoonshotAI2025} with one-shot prompting and Qwen3-14B \cite{Alibaba2025} across all subtasks. Qwen3-14B is separately fine-tuned for each dataset using QLoRA \cite{dettmers2023qlora} on the official training split. 

\paragraph{Track B.} We adopt the multilingual pretrained Transformer mBERT \cite{devlin2019bert} and Mistral-3-14B \cite{MistralAI2025}. mBERT is fully fine-tuned, while Mistral-3-14B is fine-tuned using QLoRA. Both models are trained separately on the official training split for each dataset.

Implementation details and additional baseline results are provided in our dataset papers \cite{lee2026dimabsabuildingmultilingualmultidomain, becker2026dimstancemultilingualdatasetsdimensional}.

\subsection{Task Organization}
\begin{figure}[t]
    \centering
    \includegraphics[width=\columnwidth]{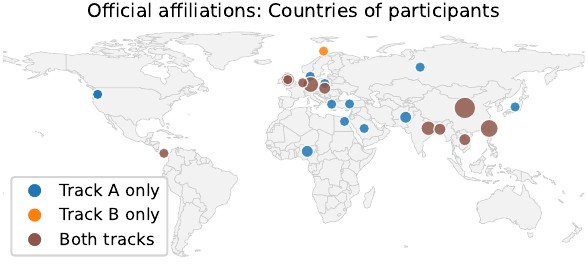}
    \caption{\textbf{Countries of official affiliations of participants.} Larger dots indicate more participants. A total of 24 countries are represented.}
        \label{fig:worldmap}
\end{figure}

We used Codabench as the competition platform and released pilot data before the shared task to help participants understand the task. We also provided a starter kit on GitHub, beginner resources, and organized a Q\&A session and a writing tutorial for junior researchers. Participants came from different parts of the world, as shown in \Cref{fig:worldmap}.

The task consisted of two phases: (1) a development phase and (2) an evaluation phase. During the development phase, the leaderboard was open, allowing up to 999 submissions per participant. During the evaluation phase, the leaderboard was closed, and each participant could submit up to four runs, with the last used for the official ranking.

%, while 23 individuals participated independently without providing an affiliation

\section{Participating Systems and Results} 

\begin{table*}[t]
\centering
\resizebox{\textwidth}{!}{
\begin{tabular}{|l l r|l l r|l l r|l l r|}
\hline
Dataset & Team & Score & Dataset & Team & Score & Dataset & Team & Score & Dataset & Team & Score \\
\hline
\multicolumn{12}{|c|}{\textbf{Subtask 1}} \\
\hline
eng-rest & LogSigma & 1.1035 & eng-lap & LogSigma & 1.2408 & jpn-hot & TeleAI & 0.5561 & jpn-fin & TeleAI & 0.6581 \\
        & Bert Kittens & 1.1812 &         & TeleAI & 1.2425 &         & PALI & 0.6237 &         & PALI & 0.7532 \\
        & Baseline(KimiK2) & 2.1461 &         & Baseline(KimiK2) & 2.1893 &         & Baseline(KimiK2) & 1.7553 &         & Baseline(KimiK2) & 1.6396 \\
        & Baseline(Qwen3-14B) & 2.6427 &         & Baseline(Qwen3-14B) & 2.8089 &         & Baseline(Qwen3-14B) & 2.2906 &         & Baseline(Qwen3-14B) & 1.8964 \\
\hline
rus-rest & PAI & 1.2190 & tat-rest & PAI & 1.5294 & ukr-rest & PAI & 1.1888 & zho-fin & HUS@NLP-VNU & 0.4841 \\
        & TeleAI & 1.2456 &         & Habib University & 1.6041 &         & TeleAI & 1.3234 &         & YangS\_team & 0.4864 \\
        & Baseline(KimiK2) & 1.7768 &         & Baseline(KimiK2) & 1.9380 &         & Baseline(KimiK2) & 1.7805 &         & Baseline(KimiK2) & 1.9652 \\
        & Baseline(Qwen3-14B) & 2.1528 &         & Baseline(Qwen3-14B) & 2.6367 &         & Baseline(Qwen3-14B) & 2.2121 &         & Baseline(Qwen3-14B) & 1.4707 \\
\hline
zho-lap & TeleAI & 0.6103 & zho-rest & ICT-NLP & 0.9256 &  &  &  &  &  &  \\
        & ICT-NLP & 0.6553 &         & TeleAI & 0.9265 &  &  &  &  &  &  \\
        & Baseline(KimiK2) & 1.6440 &         & Baseline(KimiK2) & 1.8959 &  &  &  &  &  &  \\
        & Baseline(Qwen3-14B) & 1.7706 &         & Baseline(Qwen3-14B) & 2.0073 &  &  &  &  &  &  \\
\hline

\multicolumn{12}{|c|}{\textbf{Subtask 2}} \\
\hline
eng-rest & Takoyaki & 0.7021 & eng-lap & Takoyaki & 0.6366 & jpn-hot & TeleAI & 0.5837 & rus-rest & PAI & 0.5793 \\
        & nchellwig & 0.6985 &         & PALI & 0.6242 &         & TeamLasse & 0.5694 &         & TeleAI & 0.5736 \\
        & Baseline(KimiK2) & 0.4920 &         & Baseline(KimiK2) & 0.4424 &         & Baseline(KimiK2) & 0.3464 &         & Baseline(KimiK2) & 0.4242 \\
        & Baseline(Qwen3-14B) & 0.4483 &         & Baseline(Qwen3-14B) & 0.3827 &         & Baseline(Qwen3-14B) & 0.1622 &         & Baseline(Qwen3-14B) & 0.3341 \\
\hline
tat-rest & nchellwig & 0.5119 & ukr-rest & PAI & 0.5787 & zho-lap & PALI & 0.5308 & zho-rest & PAI & 0.5638 \\
        & Takoyaki & 0.5092 &         & TeleAI & 0.5712 &         & PAI & 0.5306 &         & PALI & 0.5634 \\
        & Baseline(KimiK2) & 0.3577 &         & Baseline(KimiK2) & 0.4220 &         & Baseline(KimiK2) & 0.2494 &         & Baseline(KimiK2) & 0.3529 \\
        & Baseline(Qwen3-14B) & 0.2020 &         & Baseline(Qwen3-14B) & 0.3099 &         & Baseline(Qwen3-14B) & 0.2099 &         & Baseline(Qwen3-14B) & 0.2509 \\
\hline

\multicolumn{12}{|c|}{\textbf{Subtask 3}} \\
\hline
eng-rest & Takoyaki & 0.6514 & eng-lap & Takoyaki & 0.4227 & jpn-hot & PALI & 0.4252 & rus-rest & PAI & 0.5599 \\
        & nchellwig & 0.6403 &         & nchellwig & 0.4006 &         & Takoyaki & 0.4086 &         & PALI & 0.5496 \\
        & Baseline(KimiK2) & 0.3746 &         & Baseline(KimiK2) & 0.2795 &         & Baseline(KimiK2) & 0.1943 &         & Baseline(KimiK2) & 0.2963 \\
        & Baseline(Qwen3-14B) & 0.2673 &         & Baseline(Qwen3-14B) & 0.1529 &         & Baseline(Qwen3-14B) & 0.0400 &         & Baseline(Qwen3-14B) & 0.1682 \\
\hline
tat-rest & Takoyaki & 0.4736 & ukr-rest & PAI & 0.5437 & zho-lap & NYCU Speech Lab & 0.4824 & zho-rest & NYCU Speech Lab & 0.5521 \\
        & nchellwig & 0.4557 &         & PALI & 0.5307 &         & PALI & 0.4319 &         & PAI & 0.5360 \\
        & Baseline(KimiK2) & 0.2380 &         & Baseline(KimiK2) & 0.2971 &         & Baseline(KimiK2) & 0.1900 &         & Baseline(KimiK2) & 0.2859 \\
        & Baseline(Qwen3-14B) & 0.0954 &         & Baseline(Qwen3-14B) & 0.1641 &         & Baseline(Qwen3-14B) & 0.1124 &         & Baseline(Qwen3-14B) & 0.1605 \\
\hline
\end{tabular}}
\caption{\textbf{Track A (DimABSA) results across all subtasks.} Subtask 1 is evaluated using RMSE, while Subtasks 2 and 3 are evaluated using cF1. The top two teams and the official baseline systems are reported for each dataset.}
\label{tab:top_teams_A}
\end{table*}

\begin{table}[t]
\centering
\small
\resizebox{\columnwidth}{!}{
\begin{tabular}{|l l r|l l r|}
\hline
Dataset & Team & Score & Dataset & Team & Score \\
\hline
\multicolumn{6}{|c|}{\textbf{Subtask 1}} \\
\hline

eng-env & LogSigma & 1.4734 
        & deu-pol & LogSigma & 1.3417 \\
        & hllwan & 1.5122 
        &          & NTNU-SMIL & 1.3467 \\
        & Baseline (Mistral-3-14B) & 1.6431 
        &          & Baseline (Mistral-3-14B) & 1.5914 \\
        & Baseline (mBERT) & 2.6985 
        &          & Baseline (mBERT) & 2.3294 \\
\hline

zho-env & YangS\_team & 0.5468 
        & pcm-pol & CYUT & 1.1024 \\
        & NTNU-SMIL & 0.5561 
        &          & LogSigma & 1.1269 \\
        & Baseline (Mistral-3-14B) & 0.7403 
        &          & Baseline (Mistral-3-14B) & 1.7392 \\
        & Baseline (mBERT) & 1.2756 
        &          & Baseline (mBERT) & 3.2152 \\
\hline

swa-pol & LogSigma & 1.7959 
        &  &  &  \\
        & HUS@NLP-VNU & 1.8713 
        &  &  &  \\
        & Baseline (Mistral-3-14B) & 2.2992 
        &  &  &  \\
        & Baseline (mBERT) & 2.7835 
        &  &  &  \\
\hline

\end{tabular}
}
\caption{\textbf{Track B (DimStance) results.} Evaluation is based on RMSE. The top two teams and the official baseline systems are reported for each dataset.}
\label{tab:top_teams_B}
\end{table}

\subsection{Overview}
The task attracted over 300 participants in Track~A (DimABSA) and over 100 in Track~B (DimStance). During the development phase, 2664 submissions were made to Track A and 357 to Track B. In the evaluation phase, 177 submissions were made to Track A and 67 to Track B. While the English datasets received the most submissions, all languages had at least 20 submissions in each track.

We report results only for teams that submitted a system description paper. In total, 39 teams with 84 submissions participated in Track A, and 13 teams with 28 submissions participated in Track B, resulting in 112 submissions from 42 unique teams, including 10 teams that participated in both tracks. Participant information is listed in \Cref{tab:dimabsa-participants-info}. 

%(Track A -- English: 74, Japanese: 36, Russian: 40, Tatar: 35, Ukrainian: 32, Chinese: 42; Track B -- English: 25, German: 22, Chinese: 23, Nigerian Pidgin: 21, Swahili: 20).

\subsection{Track A: DimABSA}
Track A includes three subtasks. Subtask~1 \mbox{(DimASR)} attracted the most teams (36), followed by Subtask~2 (DimASTE) with 22 teams and Subtask~3 (DimASQP) with 20 teams. \Cref{tab:top_teams_A} presents the top two systems for each dataset across all subtasks, together with the baselines. The complete results for each subtask are reported in \Cref{tab:dimabsa-track-a-subtask-1}, \Cref{tab:dimabsa-track-a-subtask-2}, and \Cref{tab:dimabsa-track-a-subtask-3}.

The DimASR results for sentiment regression show that systems achieve lower RMSE on the Chinese and Japanese datasets, whereas the highest RMSE is observed on the low-resource Tatar dataset. DimASTE and DimASQP report results for joint structured extraction and regression. In DimASTE, systems achieve the highest cF1 on the English datasets and the lowest on the Tatar dataset. DimASQP is more difficult than DimASTE due to the additional classification of domain-dependent aspect categories. The laptop and hotel domains show a noticeable performance drop, likely due to the larger number of aspect categories and their long-tailed distribution \cite{lee2026dimabsabuildingmultilingualmultidomain}. 

\subsubsection{Best-Performing Systems}
\noindent\paragraph{Team PAI} (1st on \texttt{rus-rest}$_{\text{ST1--3}}$, \texttt{tat-rest}$_{\text{ST1}}$, \texttt{ukr-rest}$_{\text{ST1--3}}$, \texttt{zho-rest}$_{\text{ST2}}$). They propose a distributional adaptation method to align predicted VA scores with the training set distribution while preserving the inter-dimensional correlation between valence and arousal. Initial predictions are generated by Qwen3-32B \citep{yang2025qwen3technicalreport} fine-tuned with LoRA \cite{hu2021lora} and subsequently calibrated using the Sinkhorn algorithm.

\noindent\paragraph{Team TeleAI.} (1st on \texttt{jpn-hot}$_{\text{ST1--2}}$, \texttt{jpn-fin}$_{\text{ST1}}$, \texttt{zho-lap}$_{\text{ST1}}$). Their system is based on Qwen2.5-7B \cite{qwen2025qwen25} fine-tuned with LoRA. To improve generalization, they train a single multilingual, multi-domain model on all task training sets. They apply robust training, including Smooth L1 loss with R-Drop consistency, embedding-level PGD adversarial training, and post-hoc linear calibration. 

\noindent\paragraph{Team Takoyaki.} (1st on \texttt{eng-rest}$_{\text{ST2--3}}$, \texttt{eng-lap}$_{\text{ST2--3}}$, \texttt{tat-rest}$_{\text{ST3}}$). They adopt retrieval-based in-context learning, where multiple BM25 variants retrieve similar training examples for the Gemini 3.0 Pro \citep{geminiteam2023gemini} to generate quadruplet predictions. An agreement-based ensemble strategy is then applied to retain quadruplets with high agreement scores across variants. Finally, LLM-mined correction rules are applied to fix extraction and category errors. The VA scores are averaged across duplicate quadruplets after ensembling and correction.

%\texttt{eng-res} and \texttt{eng-lap} of Subtask 2, and \texttt{eng-res}, \texttt{eng-lap}, and \texttt{tat-res} of Subtask 3 \citep{yamada-semeval2026-sub246}. 

% \noindent\paragraph{Other Systems.} LogSigma placed first on both \texttt{eng-res} and \texttt{eng-lap} of Subtask 1. Their system leans on language-specific pretrained models, uncertainty-based loss weighting, and multi-seed ensembling, which suits the regression setup. HUS@NLP-VNU took first on zho-fin, using a syntax-aware dual-stream model for DimABSA and DPO-style tuning for the structured prediction subtasks. ICT-NLP won \texttt{zho-res} in Subtask 1 with a compact PLM-based approach built around multilingual, multidomain training and ensemble selection. nchellwig’s top result placed first on \texttt{tat-res} in Subtask 2, where they used Gemma 3 27B with LoRA, plus repeated generation and majority voting. NYCU Speech Lab was strongest in Chinese for Subtask 3, placing first on both \texttt{zho-lap} and \texttt{zho-res} with a practical recipe of model search, ensembling, threshold tuning, and post-processing.

\subsection{Track B: DimStance}
Track B had 13 participating teams. \Cref{tab:top_teams_B} presents the top two systems together with our baselines. The complete results are reported in \Cref{tab:dimabsa-track-b-subtask-1}. The DimASR results for stance targets show that systems achieve the lowest RMSE on the Chinese dataset, whereas the highest RMSE is observed on the low-resource Swahili dataset.

\subsubsection{Best-Performing Systems}
\noindent\paragraph{Team LogSigma} (1st on Track A: \texttt{eng-rest}$_{\text{ST1}}$, \texttt{eng-lap}$_{\text{ST1}}$; Track B: \texttt{eng-env}$_{\text{ST1}}$, \texttt{deu-pol}$_{\text{ST1}}$, \texttt{swa-pol}$_{\text{ST1}}$). They treat VA prediction as two regression tasks for valence and arousal and focus on balancing them. Instead of fixing loss weights, the model learns task-specific log-variance parameters that down-weight noisier objectives during training, allowing it to balance valence and arousal losses based on their prediction difficulty. The model uses a language-specific transformer encoder produces a shared representation, which is passed to separate regression heads. Final predictions are stabilized using a multi-seed ensemble.

%Team LogSigma participated in Subtask 1 of Tracks A and B, placing first on five datasets, including \texttt{eng-res} and \texttt{eng-lap} in Track A and \texttt{eng-env}, \texttt{deu-pol}, and \texttt{swa-pol} in Track B \citep{hikal-semeval2026-sub182}.

\noindent\paragraph{Team YangS\_team} (1st on \texttt{zho-env}$_{\text{ST1}}$). They fine-tune mDeBERTa-v3-base \citep{he2021debertadecodingenhancedbertdisentangled} with aspect-aware marker encoding to predict VA scores. The contextual representation of the aspect marker is pooled and passed to dual regression heads to jointly estimate valence and arousal, and the prediction stability is further improved through a 5-fold ensemble.

\noindent\paragraph{Team CYUT} (1st on \texttt{pcm-pol}$_{\text{ST1}}$). They introduce a geometry-informed multi-task framework to fine-tune Qwen2-7B \citep{bai2023qwentechnicalreport} with LoRA for VA regression. The framework incorporates auxiliary geometry-derived signals (polarity, intensity, quadrant, and directional prototypes), derived from the VA annotations, to stabilize training.

% Describe: hllwan, NTNU-SMIL, HUS@NLP-VNU (second places)
% \noindent\paragraph{Other Systems.} Among the other teams, hllwan placed second on \texttt{eng-env} with a system that combines dependency-based opinion extraction, DeBERTa/Qwen2.5-based feature fusion, and translation-based data augmentation for low-resource settings. NTNU-SMIL placed second on both \texttt{deu-pol} and \texttt{zho-env}, using a straightforward transformer regression setup with aspect-aware input formatting, separate prediction heads, and added calibration and ensembling. HUS@NLP-VNU took second on Swahili \texttt{swa-pol} with a syntax-aware framework supported by simple preprocessing choices such as lowercasing.

\section{Analysis and Discussion}

\paragraph{Model Architecture.} 
\Cref{fig:architectures} summarizes the architectures adopted by participating systems and shows a trend consistent with recent SemEval tasks, where systems are primarily based on pretrained transformers (e.g., RoBERTa-family models) and LLMs (e.g., Qwen), as shown in \Cref{fig:models}. Another popular approach is model ensembling. Teams constructed ensembles from models trained with different random seeds (LogSigma), cross-validation folds (YangS), and hyperparameters (ICT-NLP, 1st on \texttt{zho-rest}$_{\text{ST1}}$, Track A), as well as heterogeneous model architectures (NYCU Speech Lab, 1st on \texttt{zho-rest}$_{\text{ST3}}$ and \texttt{zho-lap}$_{\text{ST3}}$, Track A). In addition, Team HUS@NLP-VNU (1st on \texttt{zho-fin}$_{\text{ST1}}$, Track A) uses a syntax-aware Graph Convolutional Network (GCN) model. 

\paragraph{Training Techniques.} Figure~\ref{fig:trainings} summarizes the training techniques used by participating systems. Most systems rely on fine-tuning pretrained models, typically via full fine-tuning or parameter-efficient adaptation, as shown in Figure~\ref{fig:finetuning}. Beyond standard fine-tuning, some systems improve training stability, such as using Smooth L1 loss (TeleAI) and log-variance loss weighting (LogSigma). Introducing auxiliary learning signals (CYUT) and adjusting the prediction distribution (PAI) can also improve performance. Team PALI (1st on \texttt{zho-lap}$_{\text{ST2}}$, Track A) further employs per-language adapters to capture language-specific VA distributions while reducing the number of required models.

\paragraph{Prompting Strategies.}
Figure~\ref{fig:prompting} summarizes the prompting strategies adopted by participating systems, showing that instruction prompting with few-shot demonstrations is widely used. Beyond random sampling of demonstrations, in-context retrieval can improve prediction consistency by retrieving semantically similar training instances (Takoyaki). Meanwhile, Team nchellwig (1st on \texttt{tat-rest}$_{\text{ST2}}$, Track A) adopts a self-consistency strategy that executes the model multiple times with stochastic decoding and aggregates the resulting predictions via majority voting, retaining only tuples that achieve consensus to improve reliability.

%\subsection{Error Analysis}
%Across Track A, results show a clear increase in difficulty from DimASTE to DimASQP, as indicated by the decreased performance of top systems on DimASQP. Jointly extracting aspects together with opinion terms and aspect categories, while also predicting continuous VA scores, leads to substantially lower scores than aspect-level regression with only extracting opinion terms. We assume that general task difficulty is lowest for DimASR on Tracks A and B, as they require only regression to stance targets.

\section{Conclusions}
This paper presents the SemEval-2026 shared task, which extends categorical ABSA and stance detection by incorporating a dimensional \mbox{valence–arousal} representation. We organize the task into the DimABSA and DimStance tracks and introduce three subtasks, ranging from pure regression to hybrid structured extraction with regression. We also introduce a new cF1 metric that unifies categorical and continuous evaluation.

We report results on systems evaluated on our multilingual and multidomain datasets, discuss top-performing systems, and summarize key design choices. These findings highlight challenges and opportunities for advancing dimensional sentiment analysis at the aspect and stance-target levels.

\section*{Limitations}

Although DimABSA and DimStance datasets are multilingual, interpretations of valence and arousal can vary across cultures, thereby affecting cross-lingual comparability.
We mitigate this by using five native-speaker annotators per language and sample, consistent 1–9 VA scales, and shared guidelines; nonetheless, results should be interpreted as comparisons across language-community-domain settings. Expanding language coverage and testing measurement invariance are important directions for future work.

%Inter-annotator agreement is acceptable, yet consistently lower for arousal than for valence (e.g., 1.70 vs 2.20 RMSE for English). This observation is consistent with previous studies reporting lower reliability for arousal annotations \cite{buechel-hahn-2017-emobank, mohammad2018obtaining, lee2022chinese}. We average ratings across five annotators and provide explicit guidance. Downstream models should account for this uncertainty (e.g., confidence intervals, robustness checks) \citep{troiano-etal-2021-emotion}.

Some datasets (e.g., Nigerian Pidgin) include more samples of negative or positive valence, which may bias models during training and inflate performance in the majority regions of the VA space.
We document these differences and encourage explicit handling (e.g., reweighting or stratified sampling) when training or comparing models across languages \citep{10.5555/1622407.1622416}.

\section*{Ethical Considerations}

People expresses attitudes, opinions, opinions, and sentiment towards entities and their aspects in complex and nuanced ways. Further, there can be considerable person-to-person variation. It should be noted that human annotated labels capture \textbf{perceived} sentiment and attitudes, and that in several cases this may be different from the speaker's true attitudes. Nonetheless, since language is key mechanism to communicate, at an aggregate-level perceived opinions tend to correlate with actual opinions. Thus, even perceived opinions are useful at an aggregate level. However, caution must be employed when using individual inferred opinions to make decisions about individuals, especially high-stakes decisions.  

ABSA and stance detection, like many technologies, can be abused and misused. For example, it can be used to identify likes and dislikes and to manipulate people into behaviours that may not be in their best interests (e.g., purchasing products or availing services that they cannot afford or that are not particularly useful to them). This is especially concerning for vulnerable populations such as children and the elderly. We expressly forbid any commercial use of our data.

For a detailed discussion of a large number of ethical considerations associated with automatic sentiment and emotion detection, we refer the reader to \citet{mohammad-2022-ethics-sheet,mohammad-2023-best}.

\section*{Acknowledgments}

Liang-Chih Yu and Lung-Hao Lee acknowledge support from the National Science and Technology Council, Taiwan, under grants NSTC 113-2221-E-155-046-MY3 and NSTC 114-2221-E-A49-059-MY3.
\newline
Jonas Becker acknowledges the support of the Landeskriminalamt NRW.
\newline
The work of Alexander Panchenko was supported by the RSF project 25-71-30008 ``Laboratory for reliable, adaptive, and trustworthy Artificial Intelligence''. 
\newline
Ilseyar Alimova gratefully acknowledges Dina Abdullina for the Tatar data annotation and AIRI for financial support.
% @EVERYONE: Add the acknowledgements as a sentence below. 
\newline
Jan Philip Wahle, Terry Ruas, and Bela Gipp acknowledge the support of the Lower Saxony Ministry of Science and Culture, and the VW Foundation.
\newline
Shamsuddeen Hassan Muhammad acknowledges the support of Google DeepMind, whose funding made this work possible.
\newline
\bibliography{customDimStance, dimStance, DimABSA, semeval2026, custom}

\begin{thebibliography}{111}
\providecommand{\natexlab}[1]{#1}

\bibitem[{Adam et~al.(2026)Adam, Aliyu, Aji, Abubakar, and Shuaibu}]{adam-semeval2026-sub38}
Faisal~Muhammad Adam, Lukman~Jibril Aliyu, Sani Aji, Abdulhamid Abubakar, and Aliyu~Rabiu Shuaibu. 2026.
\newblock {Team faisalm3at SemEval-2026 Task 3: From Standard Regression to Distributional Alignment in Dimensional Sentiment Analysis}.
\newblock In \emph{Proceedings of the 20th International Workshop on Semantic Evaluation (SemEval-2026)}, San Diego, California. Association for Computational Linguistics.

\bibitem[{Affan et~al.(2026)Affan, Shahzad, Imam, Zulfiqar, Kumar, and Samad}]{affan-semeval2026-sub328}
Muhammad Affan, M~Hassan Shahzad, Mikaal Imam, Moiz Zulfiqar, Sandesh Kumar, and Abdul Samad. 2026.
\newblock {Habib University at SemEval-2026 Task 3: A Pipeline Approach for Dimensional Aspect-Based Sentiment Analysis}.
\newblock In \emph{Proceedings of the 20th International Workshop on Semantic Evaluation (SemEval-2026)}, San Diego, California. Association for Computational Linguistics.

\bibitem[{Alibaba(2025)}]{Alibaba2025}
Alibaba. 2025.
\newblock Qwen3 technical report.
\newblock \emph{arXiv preprint}, page arXiv:2505.09388.

\bibitem[{Allaway and McKeown(2020)}]{allaway-mckeown-2020-zero}
Emily Allaway and Kathleen McKeown. 2020.
\newblock \href {https://doi.org/10.18653/v1/2020.emnlp-main.717} {Zero-shot stance detection: A dataset and model using generalized topic representations}.
\newblock In \emph{Proceedings of the 2020 Conference on Empirical Methods in Natural Language Processing (EMNLP)}, pages 8913--8931. Association for Computational Linguistics.

\bibitem[{Alshawi et~al.(2026)Alshawi, Raj, Kudelya, and Shirnin}]{alshawi-semeval2026-sub371}
Rafif Alshawi, Amit Raj, Aleksey Kudelya, and Alexander Shirnin. 2026.
\newblock {The Classics at SemEval-2026 Task 3: Combining Transformer Models and LLM-Generated Annotations for Dimensional Aspect-Based Sentiment Analysis}.
\newblock In \emph{Proceedings of the 20th International Workshop on Semantic Evaluation (SemEval-2026)}, San Diego, California. Association for Computational Linguistics.

\bibitem[{Amol et~al.(2024)Amol, Wanzare, and Obuhuma}]{Amol_PolitikWeli}
Cynthia Amol, Lilian Wanzare, and James Obuhuma. 2024.
\newblock Politikweli: A swahili-english code-switched twitter political misinformation classification dataset.
\newblock In \emph{Speech and Language Technologies for Low-Resource Languages}, pages 3--17, Cham. Springer Nature Switzerland.

\bibitem[{Arampatzis and Arampatzis(2026)}]{arampatzis-semeval2026-sub94}
Georgios Arampatzis and Avi Arampatzis. 2026.
\newblock {DUTH at SemEval-2026 Task 3: Multilingual Transformer Models for Dimensional Stance Prediction Across Tracks}.
\newblock In \emph{Proceedings of the 20th International Workshop on Semantic Evaluation (SemEval-2026)}, San Diego, California. Association for Computational Linguistics.

\bibitem[{Ayg{\"u}n et~al.(2022)Ayg{\"u}n, Kaya, and Kaya}]{aygun2022aspect}
{\.I}rfan Ayg{\"u}n, Buket Kaya, and Mehmet Kaya. 2022.
\newblock \href {https://doi.org/10.1109/JBHI.2021.3133103} {Aspect based twitter sentiment analysis on vaccination and vaccine types in {COVID-19} pandemic with deep learning}.
\newblock \emph{IEEE Journal of Biomedical and Health Informatics}, 26(5):2360--2369.

\bibitem[{Baccianella et~al.(2010)Baccianella, Esuli, and Sebastiani}]{Baccianella2010}
Stefano Baccianella, Andrea Esuli, and Fabrizio Sebastiani. 2010.
\newblock \href {http://www.lrec-conf.org/proceedings/lrec2010/pdf/769_Paper.pdf} {Sentiwordnet 3.0: An enhanced lexical resource for sentiment analysis and opinion mining}.
\newblock In \emph{Proceedings of the 7th International Conference on Language Resources and Evaluation}, pages 2200--2204.

\bibitem[{Bai et~al.(2023)Bai, Bai, Chu, Cui, Dang, Deng, Fan, Ge, Han, Huang, Hui, Ji, Li, Lin, Lin, Liu, Liu, Lu, Lu, Ma, Men, Ren, Ren, Tan, Tan, Tu, Wang, Wang, Wang, Wu, Xu, Xu, Yang, Yang, Yang, Yang, Yao, Yu, Yuan, Yuan, Zhang, Zhang, Zhang, Zhang, Zhou, Zhou, Zhou, and Zhu}]{bai2023qwentechnicalreport}
Jinze Bai, Shuai Bai, Yunfei Chu, Zeyu Cui, Kai Dang, Xiaodong Deng, Yang Fan, Wenbin Ge, Yu~Han, Fei Huang, Binyuan Hui, Luo Ji, Mei Li, Junyang Lin, Runji Lin, Dayiheng Liu, Gao Liu, Chengqiang Lu, Keming Lu, and 29 others. 2023.
\newblock \href {https://arxiv.org/abs/2309.16609} {Qwen technical report}.
\newblock \emph{Preprint}, arXiv:2309.16609.

\bibitem[{Becker et~al.(2026)Becker, Yu, Muhammad, Wahle, Ruas, Abdulmumin, Lee, Odhiambo, Wanzare, Liu, Lin, Xu, Lin, Wang, Mukhtar, Gipp, and Mohammad}]{becker2026dimstancemultilingualdatasetsdimensional}
Jonas Becker, Liang-Chih Yu, Shamsuddeen~Hassan Muhammad, Jan~Philip Wahle, Terry Ruas, Idris Abdulmumin, Lung-Hao Lee, Nelson Odhiambo, Lilian Wanzare, Wen-Ni Liu, Tzu-Mi Lin, Zhe-Yu Xu, Ying-Lung Lin, Jin Wang, Maryam~Ibrahim Mukhtar, Bela Gipp, and Saif~M. Mohammad. 2026.
\newblock \href {https://arxiv.org/abs/2601.21483} {Dimstance: Multilingual datasets for dimensional stance analysis}.
\newblock \emph{Preprint}, arXiv:2601.21483.

\bibitem[{Bhalgat et~al.(2026)Bhalgat, Jagtap, and Phakatkar}]{bhalgat-semeval2026-sub50}
Aditya~Praful Bhalgat, Omkar~Dnyaneshwar Jagtap, and Anupama Phakatkar. 2026.
\newblock {PICT at SemEval-2026 Task 3: A Transformer-Based System for Dimensional Aspect-Aware Sentiment Regression with Weighted Layer Pooling}.
\newblock In \emph{Proceedings of the 20th International Workshop on Semantic Evaluation (SemEval-2026)}, San Diego, California. Association for Computational Linguistics.

\bibitem[{Buechel and Hahn(2017{\natexlab{a}})}]{buechel-hahn-2017-emobank}
Sven Buechel and Udo Hahn. 2017{\natexlab{a}}.
\newblock \href {aclanthology.org} {{E}mo{B}ank: {A} {C}orpus of {A}nalyze{D} {E}motions on a {D}imensional {L}evel}.
\newblock In \emph{Proceedings of the 15th Conference of the European Chapter of the Association for Computational Linguistics: Volume 2, Short Papers}, pages 572--578, Valencia, Spain. Association for Computational Linguistics.

\bibitem[{Buechel and Hahn(2017{\natexlab{b}})}]{Buechel2017}
Sven Buechel and Udo Hahn. 2017{\natexlab{b}}.
\newblock \href {https://aclanthology.org/E17-2092/} {Emobank: Studying the impact of annotation perspective and representation format on dimensional emotion analysis}.
\newblock In \emph{Proceedings of the 15th Conference of the European Chapter of the Association for Computational Linguistics}, pages 578--585.

\bibitem[{Cai et~al.(2021)Cai, Xia, and Yu}]{Cai2021}
Hongjie Cai, Rui Xia, and Jianfei Yu. 2021.
\newblock \href {https://doi.org/10.18653/v1/2021.acl-long.29} {Aspect-category-opinion-sentiment quadruple extraction with implicit aspects and opinions}.
\newblock In \emph{Proceedings of the 59th Annual Meeting of the Association for Computational Linguistics and the 11th International Joint Conference on Natural Language Processing}, pages 340--350.

\bibitem[{Cao et~al.(2026)Cao, Hoang, Toan, and Linh}]{cao-semeval2026-sub312}
An~Hai Cao, Lam~Thiet Hoang, Le~Ngoc Toan, and Ha~My Linh. 2026.
\newblock {HUS@NLP-VNU at SemEval-2026 Task 3: Dual-Stream Syntax-Aware Modeling and Direct Preference Optimization for Dimensional ABSA}.
\newblock In \emph{Proceedings of the 20th International Workshop on Semantic Evaluation (SemEval-2026)}, San Diego, California. Association for Computational Linguistics.

\bibitem[{Chawla et~al.(2002)Chawla, Bowyer, Hall, and Kegelmeyer}]{10.5555/1622407.1622416}
Nitesh~V. Chawla, Kevin~W. Bowyer, Lawrence~O. Hall, and W.~Philip Kegelmeyer. 2002.
\newblock Smote: synthetic minority over-sampling technique.
\newblock \emph{J. Artif. Int. Res.}, 16(1):321–357.

\bibitem[{Chebolu et~al.(2023)Chebolu, Dernoncourt, Lipka, and Solorio}]{chebolu-etal-2023-review}
Siva Uday~Sampreeth Chebolu, Franck Dernoncourt, Nedim Lipka, and Thamar Solorio. 2023.
\newblock \href {https://aclanthology.org/2023.ijcnlp-main.41/} {A review of datasets for aspect-based sentiment analysis}.
\newblock In \emph{Proceedings of the 13th International Joint Conference on Natural Language Processing and the 3rd Conference of the Asia-Pacific Chapter of the Association for Computational Linguistics (Volume 1: Long Papers)}, pages 611--628, Nusa Dua, Bali. Association for Computational Linguistics.

\bibitem[{Chebolu et~al.(2024)Chebolu, Dernoncourt, Lipka, and Solorio}]{Chebolu2024}
Siva Uday~Sampreeth Chebolu, Franck Dernoncourt, Nedim Lipka, and Thamar Solorio. 2024.
\newblock \href {https://aclanthology.org/2024.lrec-main.1080/} {Oats: A challenge dataset for opinion aspect target sentiment joint detection for aspect-based sentiment analysis}.
\newblock In \emph{Proceedings of the 2024 Joint International Conference on Computational Linguistics, Language Resources and Evaluation}, pages 12336--12347.

\bibitem[{Chen(2026)}]{chen-semeval2026-sub102}
Cheng Chen. 2026.
\newblock {PALI at SemEval-2026 Task 3: LoRA Fine-Tuning with Validation for DimABSA}.
\newblock In \emph{Proceedings of the 20th International Workshop on Semantic Evaluation (SemEval-2026)}, San Diego, California. Association for Computational Linguistics.

\bibitem[{Chen and Liu(2026)}]{chen-semeval2026-sub249}
Haohuan Chen and Han Liu. 2026.
\newblock {Scmhl5 at SemEval-2026 Task 3: Uncertainty-Aware Adversarial Learning for Embedding Enhancement in Dimensional Aspect-Based Sentiment Analysis}.
\newblock In \emph{Proceedings of the 20th International Workshop on Semantic Evaluation (SemEval-2026)}, San Diego, California. Association for Computational Linguistics.

\bibitem[{Dai and Lin(2026)}]{dai-semeval2026-sub238}
Songqian Dai and Wei Lin. 2026.
\newblock {ALPS-Lab at SemEval‑2026 Task 3: A Multilingual Generative LLM Approach for Dimensional Aspect Sentiment Analysis}.
\newblock In \emph{Proceedings of the 20th International Workshop on Semantic Evaluation (SemEval-2026)}, San Diego, California. Association for Computational Linguistics.

\bibitem[{{D'Aniello} et~al.(2022){D'Aniello}, Gaeta, and La~Rocca}]{d2022knowmis}
Giuseppe {D'Aniello}, Matteo Gaeta, and Ilaria La~Rocca. 2022.
\newblock Knowmis-absa: an overview and a reference model for applications of sentiment analysis and aspect-based sentiment analysis.
\newblock \emph{Artificial Intelligence Review}, 55(7):5543--5574.

\bibitem[{De~Vink et~al.(2026)De~Vink, Ventirozos, Amat-Lefort, and Han}]{devink-semeval2026-sub11}
A.J.W. De~Vink, Filippos~Karolos Ventirozos, Natalia Amat-Lefort, and Lifeng Han. 2026.
\newblock {QuadAI at SemEval-2026 Task 3: Ensemble Learning of Hybrid RoBERTa and LLMs for Dimensional Aspect-Based Sentiment Analysis}.
\newblock In \emph{Proceedings of the 20th International Workshop on Semantic Evaluation (SemEval-2026)}, San Diego, California. Association for Computational Linguistics.

\bibitem[{Dettmers et~al.(2023)Dettmers, Pagnoni, Holtzman, and Zettlemoyer}]{dettmers2023qlora}
Tim Dettmers, Artidoro Pagnoni, Ari Holtzman, and Luke Zettlemoyer. 2023.
\newblock {QLoRA}: Efficient finetuning of quantized {LLMs}.
\newblock In \emph{Advances in Neural Information Processing Systems (NeurIPS)}.

\bibitem[{Devlin et~al.(2019)Devlin, Chang, Lee, and Toutanova}]{devlin2019bert}
Jacob Devlin, Ming-Wei Chang, Kenton Lee, and Kristina Toutanova. 2019.
\newblock Bert: Pre-training of deep bidirectional transformers for language understanding.
\newblock In \emph{Proceedings of NAACL-HLT}.

\bibitem[{Dharpure and Rusnachenko(2026)}]{dharpure-semeval2026-sub315}
Harshal Dharpure and Nicolay Rusnachenko. 2026.
\newblock {hdharpure at SemEval-2026 Task 3: BERT-Based Modeling and Prediction Behavior Analysis for Multilingual Valence–Arousal Scoring}.
\newblock In \emph{Proceedings of the 20th International Workshop on Semantic Evaluation (SemEval-2026)}, San Diego, California. Association for Computational Linguistics.

\bibitem[{Ding et~al.(2025)Ding, He, Li, Zheng, He, Li, Teng, and Ji}]{ding-etal-2025-zero}
Yuzhe Ding, Kang He, Bobo Li, Li~Zheng, Haijun He, Fei Li, Chong Teng, and Donghong Ji. 2025.
\newblock \href {https://doi.org/10.18653/v1/2025.findings-acl.168} {Zero-shot conversational stance detection: Dataset and approaches}.
\newblock In \emph{Findings of the Association for Computational Linguistics: ACL 2025}, pages 3221--3235, Vienna, Austria. Association for Computational Linguistics.

\bibitem[{Frolov and Rykov(2026)}]{frolov-semeval2026-sub403}
Anatolii~Aleksanfrovich Frolov and Elisei Rykov. 2026.
\newblock {ssurface3 at SemEval-2026 Task 3: Efficient Methods for Multilingual Dimensional Aspect-Based Sentiment Analysis}.
\newblock In \emph{Proceedings of the 20th International Workshop on Semantic Evaluation (SemEval-2026)}, San Diego, California. Association for Computational Linguistics.

\bibitem[{Gazetas et~al.(2026)Gazetas, Filandrianos, Lymperaiou, Tzouveli, Voulodimos, and Stamou}]{gazetas-semeval2026-sub180}
Stavros Gazetas, George Filandrianos, Maria Lymperaiou, Paraskevi Tzouveli, Athanasios Voulodimos, and Giorgos Stamou. 2026.
\newblock {AILS-NTUA at SemEval-2026 Task 3: Efficient Dimensional Aspect-Based Sentiment Analysis}.
\newblock In \emph{Proceedings of the 20th International Workshop on Semantic Evaluation (SemEval-2026)}, San Diego, California. Association for Computational Linguistics.

\bibitem[{{Gemini Team}(2023)}]{geminiteam2023gemini}
{Gemini Team}. 2023.
\newblock Gemini: A family of highly capable multimodal models.
\newblock \emph{arXiv preprint arXiv:2312.11805}.

\bibitem[{Glandt et~al.(2021)Glandt, Khanal, Li, Caragea, and Caragea}]{GlandtKLC21}
Kyle Glandt, Sarthak Khanal, Yingjie Li, Doina Caragea, and Cornelia Caragea. 2021.
\newblock \href {https://doi.org/10.18653/v1/2021.acl-long.127} {Stance {Detection} in {COVID}-19 {Tweets}}.
\newblock In \emph{Proceedings of the 59th {Annual} {Meeting} of the {Association} for {Computational} {Linguistics} and the 11th {International} {Joint} {Conference} on {Natural} {Language} {Processing} ({Volume} 1: {Long} {Papers})}, pages 1596--1611, Online. Association for Computational Linguistics.

\bibitem[{He et~al.(2021)He, Liu, Gao, and Chen}]{he2021debertadecodingenhancedbertdisentangled}
Pengcheng He, Xiaodong Liu, Jianfeng Gao, and Weizhu Chen. 2021.
\newblock \href {https://arxiv.org/abs/2006.03654} {Deberta: Decoding-enhanced bert with disentangled attention}.
\newblock \emph{Preprint}, arXiv:2006.03654.

\bibitem[{He and Zhou(2026)}]{he-semeval2026-sub167}
Qimao He and Xiaobing Zhou. 2026.
\newblock {YNU-ABSA at SemEval-2026 Task 3: A Unified Framework for Continuous and Structured Dimensional ABSA}.
\newblock In \emph{Proceedings of the 20th International Workshop on Semantic Evaluation (SemEval-2026)}, San Diego, California. Association for Computational Linguistics.

\bibitem[{Hellwig et~al.(2026)Hellwig, Fehle, Kruschwitz, and Wolff}]{hellwig-semeval2026-sub6}
Nils~Constantin Hellwig, Jakob Fehle, Udo Kruschwitz, and Christian Wolff. 2026.
\newblock {nchellwig at SemEval-2026 Task 3: Self-Consistent Structured Generation (SCSG) for Dimensional Aspect-Based Sentiment Analysis using Large Language Models}.
\newblock In \emph{Proceedings of the 20th International Workshop on Semantic Evaluation (SemEval-2026)}, San Diego, California. Association for Computational Linguistics.

\bibitem[{Hikal et~al.(2026)Hikal, Becker, and Gipp}]{hikal-semeval2026-sub182}
Baraa Hikal, Jonas Becker, and Bela Gipp. 2026.
\newblock {LogSigma at SemEval-2026 Task 3: Uncertainty-Weighted Multitask Learning for Dimensional Aspect-Based Sentiment Analysis}.
\newblock In \emph{Proceedings of the 20th International Workshop on Semantic Evaluation (SemEval-2026)}, San Diego, California. Association for Computational Linguistics.

\bibitem[{Hou et~al.(2025)Hou, Tu, Yu, Jiang, Bah, Xu, Zhang, Yang, and Zhang}]{2025_Hou}
Linlin Hou, Wenhui Tu, Ting Yu, Ting Jiang, Mohamed Bah, Zenghui Xu, Yu~Zhang, Gaoming Yang, and Ji~Zhang. 2025.
\newblock \href {https://doi.org/10.1145/3731758} {Aspect-based sentiment analysis for covid-19: A heterogeneous graph convolutional network approach}.
\newblock \emph{ACM Transactions on Asian and Low-Resource Language Information Processing}, 24(6):1--26.

\bibitem[{Hou et~al.(2024)Hou, Li, He, Yan, Chen, and McAuley}]{hou2024bridging}
Yupeng Hou, Jiacheng Li, Zhankui He, An~Yan, Xiusi Chen, and Julian McAuley. 2024.
\newblock Bridging language and items for retrieval and recommendation.
\newblock \emph{arXiv preprint arXiv:2403.03952}.

\bibitem[{Hsieh et~al.(2026)Hsieh, Wu, and Liao}]{hsieh-semeval2026-sub225}
Hao-Chun Hsieh, Cheng-En Wu, and Yuan-Fu Liao. 2026.
\newblock {NYCU Speech Lab at SemEval-2026 Task 3: Heterogeneous Model Ensemble with Adaptive Weighted Voting for Dimensional Aspect Sentiment Quadruplet Extraction}.
\newblock In \emph{Proceedings of the 20th International Workshop on Semantic Evaluation (SemEval-2026)}, San Diego, California. Association for Computational Linguistics.

\bibitem[{Hu et~al.(2021)Hu, Shen, Wallis, Allen-Zhu, Li, Wang, and Chen}]{hu2021lora}
Edward~J. Hu, Yelong Shen, Phillip Wallis, Zeyuan Allen-Zhu, Yuanzhi Li, Shean Wang, and Weizhu Chen. 2021.
\newblock Lora: Low-rank adaptation of large language models.
\newblock \emph{arXiv preprint arXiv:2106.09685}.

\bibitem[{Hu(2026)}]{hu-semeval2026-sub140}
Shuangjin Hu. 2026.
\newblock {kirito at SemEval-2026 Task 3: Dimensional Aspect-Based Sentiment Analysis via Sentence Structure Parsing Preprocessing and Prompt-Enhanced Instruction Tuning}.
\newblock In \emph{Proceedings of the 20th International Workshop on Semantic Evaluation (SemEval-2026)}, San Diego, California. Association for Computational Linguistics.

\bibitem[{Hua et~al.(2024)Hua, Denny, Wicker, and Taskova}]{hua2024systematic}
Yan~Cathy Hua, Paul Denny, J{\"o}rg Wicker, and Katerina Taskova. 2024.
\newblock \href {https://doi.org/10.1007/s10462-024-10906-z} {A systematic review of aspect-based sentiment analysis: domains, methods, and trends}.
\newblock \emph{Artificial Intelligence Review}, 57:296.

\bibitem[{Hua et~al.(2025)Hua, Denny, Wicker, and Taskova}]{Hua2025EduRABSA}
Yan~Cathy Hua, Paul Denny, J{\"o}rg Wicker, and Katerina Taskova. 2025.
\newblock \href {https://arxiv.org/abs/2508.17008} {Edurabsa: An education review dataset for aspect-based sentiment analysis tasks}.
\newblock \emph{Preprint}, arXiv:2508.17008.

\bibitem[{Huang et~al.(2026)Huang, He, Shen, Li, and Zhang}]{huang-semeval2026-sub147}
Liyuan Huang, Jiawei He, Wutao Shen, Lin Li, and Jin Zhang. 2026.
\newblock {ICT-NLP at SemEval-2026 Task 3: Dimensional Aspect-Based Sentiment Analysis (DimABSA)}.
\newblock In \emph{Proceedings of the 20th International Workshop on Semantic Evaluation (SemEval-2026)}, San Diego, California. Association for Computational Linguistics.

\bibitem[{Iqbal et~al.(2026)Iqbal, Naswan, and Shatabda}]{iqbal-semeval2026-sub445}
Wardat~Shams Iqbal, Ruwad Naswan, and Swakkhar Shatabda. 2026.
\newblock {CLRG at SemEval-2026 Task 3: One Size Does Not Fit All: A Resource Adaptive Framework for Dimensional Sentiment Regression}.
\newblock In \emph{Proceedings of the 20th International Workshop on Semantic Evaluation (SemEval-2026)}, San Diego, California. Association for Computational Linguistics.

\bibitem[{Jones et~al.(2026)Jones, Shah, and Kahanda}]{jones-semeval2026-sub357}
Athlene Jones, Vishwaa Shah, and Indika Kahanda. 2026.
\newblock {UNF-BMI at SemEval-2026 Task 3: Research Domain Criteria-Guided Large Language Models for Dimensional Aspect-Based Sentiment Analysis}.
\newblock In \emph{Proceedings of the 20th International Workshop on Semantic Evaluation (SemEval-2026)}, San Diego, California. Association for Computational Linguistics.

\bibitem[{Kiritchenko et~al.(2016)Kiritchenko, Mohammad, and Salameh}]{Kiritchenko2016}
Svetlana Kiritchenko, Saif Mohammad, and Mohammad Salameh. 2016.
\newblock \href {https://doi.org/10.18653/v1/S16-1004} {{S}em{E}val-2016 task 7: Determining sentiment intensity of english and arabic phrases}.
\newblock In \emph{Proceedings of the10th International WorkshoponSemantic Evaluation}, pages 42--51.

\bibitem[{Kubo and Nakayama(2018)}]{Kubo2018}
Takahiro Kubo and Hiroki Nakayama. 2018.
\newblock \href {https://github.com/chakki-works/chABSA-dataset} {chabsa: Aspect-based sentiment analysis dataset}.

\bibitem[{Laschenko and Korotyk(2026)}]{laschenko-semeval2026-sub271}
Denis Laschenko and Albert Korotyk. 2026.
\newblock {SokraTUM at SemEval-2026 Task 3: A hybrid cascade of Label Distribution Learning, RAG supported generative extraction and contrastive metric learning for dimensional sentiment analysis}.
\newblock In \emph{Proceedings of the 20th International Workshop on Semantic Evaluation (SemEval-2026)}, San Diego, California. Association for Computational Linguistics.

\bibitem[{Lee et~al.(2026{\natexlab{a}})Lee, Pleva, Hladek, and Su}]{lee-semeval2026-sub214}
Chia-Yun Lee, Matus Pleva, Daniel Hladek, and Ming-Hsiang Su. 2026{\natexlab{a}}.
\newblock {SCU\_Mesclab at SemEval-2026 Task 3: An Adaptive Dual-Track Framework for Dimensional Aspect-Based Sentiment Analysis}.
\newblock In \emph{Proceedings of the 20th International Workshop on Semantic Evaluation (SemEval-2026)}, San Diego, California. Association for Computational Linguistics.

\bibitem[{Lee et~al.(2022)Lee, Li, and Yu}]{Lee2022}
Lung-Hao Lee, Jian-Hong Li, and Liang-Chih Yu. 2022.
\newblock \href {https://doi.org/10.1145/3489141} {Chinese emobank: Building valence-arousal resources for dimensional sentiment analysis}.
\newblock \emph{ACM Transactions on Asian and Low-Resource Language Information Processing}, 21(4):65.

\bibitem[{Lee et~al.(2026{\natexlab{b}})Lee, Yu, Loukashevich, Alimova, Panchenko, Lin, Xu, Zhou, Zheng, Wang, Awasthi, Becker, Wahle, Ruas, Muhammad, and Mohammad}]{lee2026dimabsabuildingmultilingualmultidomain}
Lung-Hao Lee, Liang-Chih Yu, Natalia Loukashevich, Ilseyar Alimova, Alexander Panchenko, Tzu-Mi Lin, Zhe-Yu Xu, Jian-Yu Zhou, Guangmin Zheng, Jin Wang, Sharanya Awasthi, Jonas Becker, Jan~Philip Wahle, Terry Ruas, Shamsuddeen~Hassan Muhammad, and Saif~M. Mohammad. 2026{\natexlab{b}}.
\newblock \href {https://arxiv.org/abs/2601.23022} {Dimabsa: Building multilingual and multidomain datasets for dimensional aspect-based sentiment analysis}.
\newblock \emph{Preprint}, arXiv:2601.23022.

\bibitem[{Lee et~al.(2024)Lee, Yu, Wang, and Liao}]{Lee2024}
Lung-Hao Lee, Liang-Chih Yu, Suge Wang, and Jian Liao. 2024.
\newblock \href {https://aclanthology.org/2024.sighan-1.19/} {Overview of the sighan 2024 shared task for chinese dimensional aspect-based sentiment analysis}.
\newblock In \emph{Proceedings of the 10th SIGHAN Workshop on Chinese Language Processing}, pages 165--174.

\bibitem[{Li(2026)}]{li-semeval2026-sub48}
Hongyu Li. 2026.
\newblock {SRCB at SemEval-2026 Task 3: Boosting DimASR via Contrastive LLM-Based Data Augmentation}.
\newblock In \emph{Proceedings of the 20th International Workshop on Semantic Evaluation (SemEval-2026)}, San Diego, California. Association for Computational Linguistics.

\bibitem[{Li and Yang(2026)}]{li-semeval2026-sub174}
Jinglong Li and Yang Yang. 2026.
\newblock {hllwan at SemEval-2026 Task 3: Dimensional Aspect-Based Sentiment Analysis via LLM Feature Fusion and Test-Time Adaptation}.
\newblock In \emph{Proceedings of the 20th International Workshop on Semantic Evaluation (SemEval-2026)}, San Diego, California. Association for Computational Linguistics.

\bibitem[{Li et~al.(2021)Li, Sosea, Sawant, Nair, Inkpen, and Caragea}]{LiSSN21}
Yingjie Li, Tiberiu Sosea, Aditya Sawant, Ajith~Jayaraman Nair, Diana Inkpen, and Cornelia Caragea. 2021.
\newblock \href {https://doi.org/10.18653/v1/2021.findings-acl.208} {P-{Stance}: {A} {Large} {Dataset} for {Stance} {Detection} in {Political} {Domain}}.
\newblock In \emph{Findings of the {Association} for {Computational} {Linguistics}: {ACL}-{IJCNLP} 2021}, pages 2355--2365, Online. Association for Computational Linguistics.

\bibitem[{Lin et~al.(2026)Lin, Lo, Sun, Guo, Hao, Tsai, and Chen}]{lin-semeval2026-sub3}
Siang-Ting Lin, Tien-Hong Lo, Yun-Ting Sun, Jhih-Rong Guo, Tung-Yen Hao, Fong-Chun Tsai, and Berlin Chen. 2026.
\newblock {NTNU-SMIL at SemEval-2026 Task 3: Logistic-Loss Regression with Same-Language Transfer for Valence--Arousal Stance Prediction in Dimensional Stance Analysis (DimStance)}.
\newblock In \emph{Proceedings of the 20th International Workshop on Semantic Evaluation (SemEval-2026)}, San Diego, California. Association for Computational Linguistics.

\bibitem[{Marreddy et~al.(2025)Marreddy, Oota, Chinni, Gupta, and Flek}]{marreddy-etal-2025-usdc}
Mounika Marreddy, Subba~Reddy Oota, Venkata~Charan Chinni, Manish Gupta, and Lucie Flek. 2025.
\newblock \href {https://doi.org/10.18653/v1/2025.findings-acl.1216} {{USDC}: A dataset of $\underline{U}$ser $\underline{S}$tance and $\underline{D}$ogmatism in long $\underline{C}$onversations}.
\newblock In \emph{Findings of the Association for Computational Linguistics: ACL 2025}, pages 23715--23759, Vienna, Austria. Association for Computational Linguistics.

\bibitem[{MistralAI(2025)}]{MistralAI2025}
MistralAI. 2025.
\newblock Introducing mistral 3.
\newblock \emph{mistral.ai}, pages Accessed: 2025--12--31.

\bibitem[{Modi and Szymanski(2026)}]{modi-semeval2026-sub444}
Mohammed~Shahid Modi and Boleslaw Szymanski. 2026.
\newblock {RPI Team at SemEval-2026 Task 3: An LLM-Encoder Ensemble for Coarse-to-Fine Valence-Arousal Sentiment Prediction}.
\newblock In \emph{Proceedings of the 20th International Workshop on Semantic Evaluation (SemEval-2026)}, San Diego, California. Association for Computational Linguistics.

\bibitem[{Mohammad(2018)}]{mohammad2018obtaining}
Saif Mohammad. 2018.
\newblock Obtaining reliable human ratings of valence, arousal, and dominance for 20,000 english words.
\newblock In \emph{Proceedings of the 56th Annual Meeting of the Association for Computational Linguistics (volume 1: Long papers)}, pages 174--184.

\bibitem[{Mohammad(2023)}]{mohammad-2023-best}
Saif Mohammad. 2023.
\newblock \href {https://doi.org/10.18653/v1/2023.findings-eacl.136} {Best practices in the creation and use of emotion lexicons}.
\newblock In \emph{Findings of the Association for Computational Linguistics: EACL 2023}, pages 1825--1836, Dubrovnik, Croatia. Association for Computational Linguistics.

\bibitem[{Mohammad and Bravo-Marquez(2017)}]{mohammad-bravo-marquez-2017-emotion}
Saif Mohammad and Felipe Bravo-Marquez. 2017.
\newblock \href {https://doi.org/10.18653/v1/S17-1007} {Emotion intensities in tweets}.
\newblock In \emph{Proceedings of the 6th Joint Conference on Lexical and Computational Semantics}, pages 65--77.

\bibitem[{Mohammad et~al.(2018)Mohammad, Bravo-Marquez, Salameh, and Kiritchenko}]{Mohammad2018semeval}
Saif Mohammad, Felipe Bravo-Marquez, Mohammad Salameh, and Svetlana Kiritchenko. 2018.
\newblock \href {https://doi.org/10.18653/v1/S18-1001} {{S}em{E}val-2018 task 1: Affect in tweets}.
\newblock In \emph{Proceedings of the 12th International Workshop on Semantic Evaluation}, pages 1--17.

\bibitem[{Mohammad et~al.(2016)Mohammad, Kiritchenko, Sobhani, Zhu, and Cherry}]{MohammadKSZ16a}
Saif Mohammad, Svetlana Kiritchenko, Parinaz Sobhani, Xiaodan Zhu, and Colin Cherry. 2016.
\newblock \href {https://aclanthology.org/L16-1623/} {A {Dataset} for {Detecting} {Stance} in {Tweets}}.
\newblock In \emph{Proceedings of the {Tenth} {International} {Conference} on {Language} {Resources} and {Evaluation} ({LREC}'16)}, pages 3945--3952, Portorož, Slovenia. European Language Resources Association (ELRA).

\bibitem[{Mohammad(2022)}]{mohammad-2022-ethics-sheet}
Saif~M. Mohammad. 2022.
\newblock \href {https://doi.org/10.1162/coli_a_00433} {Ethics sheet for automatic emotion recognition and sentiment analysis}.
\newblock \emph{Computational Linguistics}, 48(2):239--278.

\bibitem[{Mohammad(2025)}]{mohammad2025nrcvad}
Saif~M. Mohammad. 2025.
\newblock \href {https://arxiv.org/abs/2503.23547} {{NRC VAD Lexicon v2: Norms for Valence, Arousal, and Dominance for over 55k English Terms}}.
\newblock \emph{arXiv}, abs/2503.23547.
\newblock ArXiv:2503.23547.

\bibitem[{Mohammad et~al.(2017)Mohammad, Sobhani, and Kiritchenko}]{mohammad2017stance}
Saif~M. Mohammad, Parinaz Sobhani, and Svetlana Kiritchenko. 2017.
\newblock \href {https://doi.org/10.1145/3003433} {Stance and sentiment in tweets}.
\newblock \emph{ACM Transactions on Internet Technology}, 17(3):26:1--26:23.

\bibitem[{MoonshotAI(2025)}]{MoonshotAI2025}
MoonshotAI. 2025.
\newblock Kimi k2: Open agentic intelligence.
\newblock \emph{arXiv preprint}, page arXiv:2507.20534.

\bibitem[{Muhammad et~al.(2023)Muhammad, Abdulmumin, Ayele, Ousidhoum, Adelani, Yimam, Ahmad, Beloucif, Mohammad, Ruder, Hourrane, Brazdil, Jorge, Ali, David, Osei, Shehu~Bello, Ibrahim, Gwadabe, Rutunda, Belay, Messelle, Balcha, Chala, Gebremichael, Opoku, and Arthur}]{muhammad-etal-2023-afrisenti}
Shamsuddeen~Hassan Muhammad, Idris Abdulmumin, Abinew~Ali Ayele, Nedjma Ousidhoum, David~Ifeoluwa Adelani, Seid~Muhie Yimam, Ibrahim~Sa'id Ahmad, Meriem Beloucif, Saif~M. Mohammad, Sebastian Ruder, Oumaima Hourrane, Pavel Brazdil, Alipio Jorge, Felermino D{\'a}rio M{\'a}rio~Ant{\'o}nio Ali, Davis David, Salomey Osei, Bello Shehu~Bello, Falalu Ibrahim, Tajuddeen Gwadabe, and 8 others. 2023.
\newblock \href {https://doi.org/10.18653/v1/2023.emnlp-main.862} {{A}fri{S}enti: A {T}witter sentiment analysis benchmark for {A}frican languages}.
\newblock In \emph{Proceedings of the 2023 Conference on Empirical Methods in Natural Language Processing}, pages 13968--13981, Singapore. Association for Computational Linguistics.

\bibitem[{Muhammad et~al.(2025)Muhammad, Ousidhoum, Abdulmumin, Wahle, Ruas, Beloucif, de~Kock, Surange, Teodorescu, Ahmad, Adelani, Aji, Ali, Alimova, Araujo, Babakov, Baes, Bucur, Bukula, Cao, Tufi{\~n}o, Chevi, Chukwuneke, Ciobotaru, Dementieva, Gadanya, Geislinger, Gipp, Hourrane, Ignat, Lawan, Mabuya, Mahendra, Marivate, Panchenko, Piper, Ferreira, Protasov, Rutunda, Shrivastava, Udrea, Wanzare, Wu, Wunderlich, Zhafran, Zhang, Zhou, and Mohammad}]{muhammad-etal-2025-brighter}
Shamsuddeen~Hassan Muhammad, Nedjma Ousidhoum, Idris Abdulmumin, Jan~Philip Wahle, Terry Ruas, Meriem Beloucif, Christine de~Kock, Nirmal Surange, Daniela Teodorescu, Ibrahim~Said Ahmad, David~Ifeoluwa Adelani, Alham~Fikri Aji, Felermino D. M.~A. Ali, Ilseyar Alimova, Vladimir Araujo, Nikolay Babakov, Naomi Baes, Ana-Maria Bucur, Andiswa Bukula, and 29 others. 2025.
\newblock \href {https://doi.org/10.18653/v1/2025.acl-long.436} {{BRIGHTER}: {BRI}dging the gap in human-annotated textual emotion recognition datasets for 28 languages}.
\newblock In \emph{Proceedings of the 63rd Annual Meeting of the Association for Computational Linguistics (Volume 1: Long Papers)}, pages 8895--8916, Vienna, Austria. Association for Computational Linguistics.

\bibitem[{Ombui(2022)}]{Ombui}
Edward Ombui. 2022.
\newblock \href {https://www.kaggle.com/datasets/edwardombui/hatespeech-kenya} {{HateSpeech}\_kenya}.

\bibitem[{Peng et~al.(2020)Peng, Xu, Bing, Huang, Lu, and Si}]{Peng2020}
Haiyun Peng, Lu~Xu, Lidong Bing, Fei Huang, Wei Lu, and Luo Si. 2020.
\newblock \href {https://doi.org/10.1609/AAAI.V34I05.6383} {Knowing what, how and why: A near complete solution for aspect-based sentiment analysis}.
\newblock \emph{Proceedings of the 34th AAAI Conference on Artificial Intelligence}, 34(5):8600--8607.

\bibitem[{Pontiki et~al.(2016)Pontiki, Galanis, Papageorgiou, Androutsopoulos, Manandhar, Al-Smadi, Al-Ayyoub, Zhao, Qin, De~Clercq et~al.}]{pontiki2016semeval}
Maria Pontiki, Dimitrios Galanis, Harris Papageorgiou, Ion Androutsopoulos, Suresh Manandhar, Mohammad Al-Smadi, Mahmoud Al-Ayyoub, Yanyan Zhao, Bing Qin, Orph{\'e}e De~Clercq, and 1 others. 2016.
\newblock {S}em{E}val-2016 task 5: Aspect based sentiment analysis.
\newblock In \emph{Proceedings of the 10th International Workshop on Semantic Evaluation}, pages 19--30.

\bibitem[{Pontiki et~al.(2015)Pontiki, Galanis, Papageorgiou, Manandhar, and Androutsopoulos}]{pontiki-etal-2015-semeval}
Maria Pontiki, Dimitris Galanis, Haris Papageorgiou, Suresh Manandhar, and Ion Androutsopoulos. 2015.
\newblock \href {https://doi.org/10.18653/v1/S15-2082} {{S}em{E}val-2015 task 12: Aspect based sentiment analysis}.
\newblock In \emph{Proceedings of the 9th International Workshop on Semantic Evaluation}, pages 486--495.

\bibitem[{Pontiki et~al.(2014)Pontiki, Galanis, Pavlopoulos, Papageorgiou, Androutsopoulos, and Manandhar}]{pontiki-etal-2014-semeval}
Maria Pontiki, Dimitris Galanis, John Pavlopoulos, Harris Papageorgiou, Ion Androutsopoulos, and Suresh Manandhar. 2014.
\newblock \href {https://doi.org/10.3115/v1/S14-2004} {{S}em{E}val-2014 task 4: Aspect based sentiment analysis}.
\newblock In \emph{Proceedings of the 8th International Workshop on Semantic Evaluation}, pages 27--35.

\bibitem[{Preo{\c{t}}iuc-Pietro et~al.(2016)Preo{\c{t}}iuc-Pietro, Schwartz, Park, Eichstaedt, Kern, Ungar, and Shulman}]{preotiuc-pietro-etal-2016-modelling}
Daniel Preo{\c{t}}iuc-Pietro, H.~Andrew Schwartz, Gregory Park, Johannes Eichstaedt, Margaret Kern, Lyle Ungar, and Elisabeth Shulman. 2016.
\newblock \href {https://doi.org/10.18653/v1/W16-0404} {Modelling valence and arousal in {F}acebook posts}.
\newblock In \emph{Proceedings of the 7th Workshop on Computational Approaches to Subjectivity, Sentiment and Social Media Analysis}, pages 9--15, San Diego, California. Association for Computational Linguistics.

\bibitem[{{Qwen Team}(2025)}]{qwen2025qwen25}
{Qwen Team}. 2025.
\newblock Qwen2.5 technical report.
\newblock \emph{arXiv preprint arXiv:2412.15115}.

\bibitem[{Riewe-Perła and Filipowska(2026)}]{riewepera-semeval2026-sub59}
Oskar Riewe-Perła and Agata Filipowska. 2026.
\newblock {PUEB-DimASR at SemEval-2026 Task 3: Escaping the Mean Regression Trap with Graph-Enhanced Transformers for Dimensional Aspect-Based Sentiment Regression}.
\newblock In \emph{Proceedings of the 20th International Workshop on Semantic Evaluation (SemEval-2026)}, San Diego, California. Association for Computational Linguistics.

\bibitem[{Rosenthal et~al.(2015)Rosenthal, Nakov, Kiritchenko, Mohammad, Ritter, and Stoyanov}]{Rosenthal2015}
Sara Rosenthal, Preslav Nakov, Svetlana Kiritchenko, Saif Mohammad, Alan Ritter, and Veselin Stoyanov. 2015.
\newblock \href {https://doi.org/10.18653/v1/S15-2078} {{S}em{E}val-2015 task 10: Sentiment analysis in twitter}.
\newblock In \emph{Proceedings of the 9th International WorkshoponSemantic Evaluation}, pages 451--463.

\bibitem[{Ruan et~al.(2026)Ruan, Yang, Chen, Dai, and Mao}]{ruan-semeval2026-sub215}
Zhihao Ruan, Kaifeng Yang, Cheng Chen, Wenwen Dai, and Wenjia Mao. 2026.
\newblock {PAI at SemEval-2026 Task 3: An LLM and Data Redistribution Adaptation-Based Predictive Strategy for Valence-Arousal Scores}.
\newblock In \emph{Proceedings of the 20th International Workshop on Semantic Evaluation (SemEval-2026)}, San Diego, California. Association for Computational Linguistics.

\bibitem[{Russell(1980)}]{russell1980circumplex}
James~A Russell. 1980.
\newblock A circumplex model of affect.
\newblock \emph{Journal of personality and social psychology}, 39(6):1161--1178.

\bibitem[{Russell(2003)}]{russell2003core}
James~A Russell. 2003.
\newblock Core affect and the psychological construction of emotion.
\newblock \emph{Psychological review}, 110(1):145.

\bibitem[{Rynowiecki and Van Der~Goot(2026)}]{rynowiecki-semeval2026-sub199}
Michal Rynowiecki and Rob Van Der~Goot. 2026.
\newblock {Team BOBW (Best Of Both Worlds) at SemEval-2026 Task 3: Modular Cross-Attention Encoders for Dimensional Aspect-Based Sentiment Analysis}.
\newblock In \emph{Proceedings of the 20th International Workshop on Semantic Evaluation (SemEval-2026)}, San Diego, California. Association for Computational Linguistics.

\bibitem[{S and S(2026)}]{s-semeval2026-sub125}
Jithu~Morrison S and Abisha~Rose S. 2026.
\newblock {Pixel Phantoms at SemEval-2026 Task 3: Language-Specific Transformer Regression for Dimensional Aspect-Based Sentiment Analysis}.
\newblock In \emph{Proceedings of the 20th International Workshop on Semantic Evaluation (SemEval-2026)}, San Diego, California. Association for Computational Linguistics.

\bibitem[{Socher et~al.(2013)Socher, Perelygin, Wu, Chuang, Manning, Ng, and Potts}]{Socher2013}
Richard Socher, Alex Perelygin, Jean Wu, Jason Chuang, Christopher~D. Manning, Andrew Ng, and Christopher Potts. 2013.
\newblock \href {https://aclanthology.org/D13-1170/} {Recursive deep models for semantic compositionality over a sentiment treebank}.
\newblock In \emph{Proceedings of 2013 Conference on Empirical Methods in Natural Language Processing}, pages 1631--1642.

\bibitem[{Strothe et~al.(2026)Strothe, Sha~Kolli, and Diesner}]{strothe-semeval2026-sub306}
Lasse Strothe, Shaghayegh Sha~Kolli, and Jana Diesner. 2026.
\newblock {TeamLasse at SemEval-2026 Task 3: A Hybrid Generative-Discriminative Framework for Dimensional Aspect-Based Sentiment Analysis}.
\newblock In \emph{Proceedings of the 20th International Workshop on Semantic Evaluation (SemEval-2026)}, San Diego, California. Association for Computational Linguistics.

\bibitem[{Sukhodolsky et~al.(2026)Sukhodolsky, Salimgareev, and Ianshina}]{sukhodolsky-semeval2026-sub418}
Arseny Sukhodolsky, Ruslan Salimgareev, and Tatiana Ianshina. 2026.
\newblock {BertKittens at SemEval-2026 Task 3: Multi-Domain Aspect Sentiment with BERT/DeBERTa Ensembles for VA Regression and Aspect–Opinion–VA Triplets}.
\newblock In \emph{Proceedings of the 20th International Workshop on Semantic Evaluation (SemEval-2026)}, San Diego, California. Association for Computational Linguistics.

\bibitem[{Taboada et~al.(2011)Taboada, Brooke, Tofiloski, Voll, and Stede}]{Taboada2011}
Maite Taboada, Julian Brooke, Milan Tofiloski, Kimberly Voll, and Manfred Stede. 2011.
\newblock \href {https://doi.org/10.1162/COLI_a_00049} {Lexicon-based methods for sentiment analysis}.
\newblock \emph{Computational Linguistics}, 37(2):267--307.

\bibitem[{Thelwall et~al.(2012)Thelwall, Buckley, and Paltoglou}]{Thelwall2012}
Mike Thelwall, Kevan Buckley, and Georgios Paltoglou. 2012.
\newblock \href {https://doi.org/10.1002/asi.21662} {Sentiment strength detection for the social web}.
\newblock \emph{Journal of the Association for Information Science and Technology}, 63(1):163--173.

\bibitem[{Thenuwara et~al.(2026)Thenuwara, Mel, and De~Silva}]{thenuwara-semeval2026-sub250}
Vishal Thenuwara, Widanalage Mario Yomal~De Mel, and Nisansa De~Silva. 2026.
\newblock {Team VYN at SemEval-2026 Task 3: Dimensional Aspect-Based Sentiment Analysis}.
\newblock In \emph{Proceedings of the 20th International Workshop on Semantic Evaluation (SemEval-2026)}, San Diego, California. Association for Computational Linguistics.

\bibitem[{Vamvas and Sennrich(2020)}]{VamvasS20b}
Jannis Vamvas and Rico Sennrich. 2020.
\newblock \href {https://doi.org/10.48550/arXiv.2003.08385} {X-{Stance}: {A} {Multilingual} {Multi}-{Target} {Dataset} for {Stance} {Detection}}.
\newblock \emph{arXiv preprint}.
\newblock ArXiv:2003.08385 [cs].

\bibitem[{Wu et~al.(2025)Wu, Ma, Liu, Zhang, Deng, Li, Chen, Zhang, Xue, and Plank}]{Wu2025}
ChengYan Wu, Bolei Ma, Yihong Liu, Zheyu Zhang, Ningyuan Deng, Yanshu Li, Baolan Chen, Yi~Zhang, Yun Xue, and Barbara Plank. 2025.
\newblock \href {https://doi.org/10.18653/v1/2025.emnlp-main.128} {M-absa: A multilingual dataset for aspect-based sentiment analysis}.
\newblock In \emph{Proceedings of the 2025 Conference on Empirical Methods in Natural Language Processing}, pages 2530--2557.

\bibitem[{Wu et~al.(2026{\natexlab{a}})Wu, Chen, and Jian}]{wu-semeval2026-sub26}
Shih-Hung Wu, Xian-Yan Chen, and Yi-Min Jian. 2026{\natexlab{a}}.
\newblock {CYUT at SemEval-2026 Task 3: Multi-Task Dimensional Aspect Sentiment Regression by Fine-tuning Pretrained Models in a VA Space with Seven Emotions Directional Prototypes}.
\newblock In \emph{Proceedings of the 20th International Workshop on Semantic Evaluation (SemEval-2026)}, San Diego, California. Association for Computational Linguistics.

\bibitem[{Wu et~al.(2026{\natexlab{b}})Wu, Rusnachenko, and Liang}]{wu-semeval2026-sub270}
Tong Wu, Nicolay Rusnachenko, and Huizhi(elly) Liang. 2026{\natexlab{b}}.
\newblock {NCL-BU at SemEval-2026 Task 3: Fine-tuning XLM-RoBERTa for Multilingual Dimensional Sentiment Regression}.
\newblock In \emph{Proceedings of the 20th International Workshop on Semantic Evaluation (SemEval-2026)}, San Diego, California. Association for Computational Linguistics.

\bibitem[{Xu et~al.(2020)Xu, Li, Lu, and Bing}]{Xu2020}
Lu~Xu, Hao Li, Wei Lu, and Lidong Bing. 2020.
\newblock \href {https://doi.org/10.18653/v1/2020.emnlp-main.183} {Position-aware tagging for aspect sentiment triplet extraction}.
\newblock In \emph{Proceedings of the 2020 Conference on Empirical Methods in Natural Language Processing}, pages 2339--2349.

\bibitem[{Yamada et~al.(2026)Yamada, Takase, and Kohita}]{yamada-semeval2026-sub246}
Kosuke Yamada, Sho Takase, and Ryosuke Kohita. 2026.
\newblock {Takoyaki at SemEval-2026 Task 3: Ensembling LLM Predictions using Demonstration Retrieval for Dimensional Aspect-based Sentiment Analysis}.
\newblock In \emph{Proceedings of the 20th International Workshop on Semantic Evaluation (SemEval-2026)}, San Diego, California. Association for Computational Linguistics.

\bibitem[{Yang et~al.(2025)Yang, Li, Yang, Zhang, Hui, Zheng, Yu, Gao, Huang, Lv, Zheng, Liu, Zhou, Huang, Hu, Ge, Wei, Lin, Tang, Yang, Tu, Zhang, Yang, Yang, Zhou, Zhou, Lin, Dang, Bao, Yang, Yu, Deng, Li, Xue, Li, Zhang, Wang, Zhu, Men, Gao, Liu, Luo, Li, Tang, Yin, Ren, Wang, Zhang, Ren, Fan, Su, Zhang, Zhang, Wan, Liu, Wang, Cui, Zhang, Zhou, and Qiu}]{yang2025qwen3technicalreport}
An~Yang, Anfeng Li, Baosong Yang, Beichen Zhang, Binyuan Hui, Bo~Zheng, Bowen Yu, Chang Gao, Chengen Huang, Chenxu Lv, Chujie Zheng, Dayiheng Liu, Fan Zhou, Fei Huang, Feng Hu, Hao Ge, Haoran Wei, Huan Lin, Jialong Tang, and 41 others. 2025.
\newblock \href {https://arxiv.org/abs/2505.09388} {Qwen3 technical report}.
\newblock \emph{Preprint}, arXiv:2505.09388.

\bibitem[{Yang et~al.(2026)Yang, Hu, and Li}]{yang-semeval2026-sub218}
Liu Yang, Gang Hu, and Jing Li. 2026.
\newblock {looploop at SemEval-2026 Task 3: A Dimensional Aspect-Based Sentiment System with DeBERTa Regression and Qwen3 Instruction Fine-Tuning}.
\newblock In \emph{Proceedings of the 20th International Workshop on Semantic Evaluation (SemEval-2026)}, San Diego, California. Association for Computational Linguistics.

\bibitem[{Yang and Yang(2026)}]{yang-semeval2026-sub221}
Tsung-Hsien Yang and Shu-Fei Yang. 2026.
\newblock {YangS\_team at SemEval-2026 Task 3: Transformer-Based Aspect-Aware Regression for Dimensional Sentiment and Stance Analysis}.
\newblock In \emph{Proceedings of the 20th International Workshop on Semantic Evaluation (SemEval-2026)}, San Diego, California. Association for Computational Linguistics.

\bibitem[{Yu and Liu(2026)}]{yu-semeval2026-sub17}
Kuanlin Yu and Wen-Ni Liu. 2026.
\newblock {kevinyu66 at SemEval-2026 Task 3: A Retrieval-Augmented LLM System for Aspect–Opinion Triplet Extraction}.
\newblock In \emph{Proceedings of the 20th International Workshop on Semantic Evaluation (SemEval-2026)}, San Diego, California. Association for Computational Linguistics.

\bibitem[{Yu et~al.(2016)Yu, Lee, Hao, Wang, He, Hu, Lai, and Zhang}]{Yu2016}
Liang-Chih Yu, Lung-Hao Lee, Shuai Hao, Jin Wang, Yunchao He, Jun Hu, K.~Robert Lai, and Xuejie Zhang. 2016.
\newblock \href {https://doi.org/10.18653/v1/N16-1066} {Building chinese affective resources in valence-arousal dimensions}.
\newblock In \emph{Proceedings of the 15th Annual Conference of the North American Chapter of the Association for Computational Linguistics: Human Language Technologies}, pages 540--545.

\bibitem[{Zhang et~al.(2021)Zhang, Deng, Li, Yuan, Bing, and Lam}]{Zhang2021}
Wenxuan Zhang, Yang Deng, Xin Li, Yifei Yuan, Lidong Bing, and Wai Lam. 2021.
\newblock \href {https://doi.org/10.18653/v1/2021.emnlp-main.726} {Aspect sentiment quad prediction as paraphrase generation}.
\newblock In \emph{Proceedings of the 2021 Conference on Empirical Methods in Natural Language Processing}, pages 9209--9219.

\bibitem[{Zhang et~al.(2023)Zhang, Li, Deng, Bing, and Lam}]{Zhang2023}
Wenxuan Zhang, Xin Li, Yang Deng, Lidong Bing, and Wai Lam. 2023.
\newblock \href {https://doi.org/10.1016/j.elerap.2024.101436} {A survey on aspect-based sentiment analysis: tasks, methods, and challenges}.
\newblock \emph{IEEE Transactions on Knowledge and Data Engineering}, 35(11019--11038):101436.

\bibitem[{Zhang et~al.(2025)Zhang, Zhang, Cheng, and Xu}]{zhang-etal-2025-mad}
ZhaoDan Zhang, Jin Zhang, Xueqi Cheng, and Hui Xu. 2025.
\newblock \href {https://doi.org/10.18653/v1/2025.emnlp-main.30} {{T}-{MAD}: Target-driven multimodal alignment for stance detection}.
\newblock In \emph{Proceedings of the 2025 Conference on Empirical Methods in Natural Language Processing}, pages 580--595, Suzhou, China. Association for Computational Linguistics.

\bibitem[{Zhao and Caragea(2024)}]{zhao-caragea-2024-ez}
Chenye Zhao and Cornelia Caragea. 2024.
\newblock \href {https://doi.org/10.18653/v1/2024.acl-long.838} {{EZ}-{STANCE}: A large dataset for {E}nglish zero-shot stance detection}.
\newblock In \emph{Proceedings of the 62nd Annual Meeting of the Association for Computational Linguistics (Volume 1: Long Papers)}, pages 15697--15714, Bangkok, Thailand. Association for Computational Linguistics.

\bibitem[{Zhao et~al.(2023)Zhao, Li, and Caragea}]{zhao-etal-2023-c}
Chenye Zhao, Yingjie Li, and Cornelia Caragea. 2023.
\newblock \href {https://doi.org/10.18653/v1/2023.acl-long.747} {{C}-{STANCE}: A large dataset for {C}hinese zero-shot stance detection}.
\newblock In \emph{Proceedings of the 61st Annual Meeting of the Association for Computational Linguistics (Volume 1: Long Papers)}, pages 13369--13385, Toronto, Canada. Association for Computational Linguistics.

\bibitem[{Zhou et~al.(2025)Zhou, Peng, Luebke, Ha{\ss}ler, Haim, Mohammad, and Plank}]{zhou-etal-2025-media-frames-stance}
Shijia Zhou, Siyao Peng, Simon~M. Luebke, J{\"o}rg Ha{\ss}ler, Mario Haim, Saif~M. Mohammad, and Barbara Plank. 2025.
\newblock \href {https://doi.org/10.18653/v1/2025.findings-emnlp.286} {What media frames reveal about stance: A dataset and study about memes in climate change discourse}.
\newblock In \emph{Findings of the Association for Computational Linguistics: EMNLP 2025}, pages 5337--5356.

\bibitem[{Zhou et~al.(2026{\natexlab{a}})Zhou, Wang, Wang, Bao, Fang, Song, Li, and Li}]{zhou-semeval2026-sub262}
Yan Zhou, Wangshicheng~Shicheng Wang, Shiquan Wang, Mengjiao Bao, Ruiyu Fang, Shuangyong Song, Yongxiang Li, and Xuelong Li. 2026{\natexlab{a}}.
\newblock {TeleAI at SemEval-2026 Task 3: Large Language Models for Dimensional Aspect-Based Sentiment Analysis}.
\newblock In \emph{Proceedings of the 20th International Workshop on Semantic Evaluation (SemEval-2026)}, San Diego, California. Association for Computational Linguistics.

\bibitem[{Zhou et~al.(2026{\natexlab{b}})Zhou, He, and Bai}]{zhou-semeval2026-sub284}
Ziang Zhou, Xiangmei He, and Chenhongyi Bai. 2026{\natexlab{b}}.
\newblock {SCUZANE at SemEval-2026 Task 3: Dimension Aspect-based Sentiment Analysis with Supervised Contrastive Regression and R-Drop Regularization}.
\newblock In \emph{Proceedings of the 20th International Workshop on Semantic Evaluation (SemEval-2026)}, San Diego, California. Association for Computational Linguistics.

\bibitem[{Zotova et~al.(2020)Zotova, Agerri, Nuñez, and Rigau}]{ZotovaANR20}
Elena Zotova, Rodrigo Agerri, Manuel Nuñez, and German Rigau. 2020.
\newblock \href {https://aclanthology.org/2020.lrec-1.171/} {Multilingual {Stance} {Detection} in {Tweets}: {The} {Catalonia} {Independence} {Corpus}}.
\newblock In \emph{Proceedings of the {Twelfth} {Language} {Resources} and {Evaluation} {Conference}}, pages 1368--1375, Marseille, France. European Language Resources Association.

\end{thebibliography}

\appendix
\onecolumn

\section{Aspect Category List}
\begin{itemize}[leftmargin=*]
    \item Laptop
%\vspace{-8pt}
\begin{table}[H]
\centering
\small
\setlength{\tabcolsep}{4pt}
\renewcommand{\arraystretch}{0.4}
    \begin{subtable}{1.0\textwidth}
        \centering
        \renewcommand{\arraystretch}{1.5} 
        \begin{tabularx}{\textwidth}{|>{\arraybackslash}X|}
            \hline
            \textbf{Entity Labels} \\ \hline
            LAPTOP, DISPLAY, KEYBOARD, MOUSE, MOTHERBOARD, CPU, FANS\_COOLING, PORTS, MEMORY, POWER\_SUPPLY, OPTICAL\_DRIVES, BATTERY, GRAPHICS, HARD\_DISK, MULTIMEDIA\_DEVICES, HARDWARE, SOFTWARE, OS, WARRANTY, SHIPPING, SUPPORT, COMPANY \\ \hline
            \textbf{Attribute Labels} \\ \hline
           GENERAL, PRICE, QUALITY, DESIGN\_FEATURES, OPERATION\_PERFORMANCE, USABILITY, PORTABILITY, CONNECTIVITY, MISCELLANEOUS \\ \hline
        \end{tabularx}
        % \caption{Laptop}
    \end{subtable}
%\caption*{Laptop}
\label{tab:Label Category}
\end{table}
% \vspace{-10pt}
    \item Restaurant
%\vspace{-8pt}
\begin{table}[H]
\centering
\small
\setlength{\tabcolsep}{4pt}
\renewcommand{\arraystretch}{0.4}
\begin{subtable}[t]{1.0\textwidth}
        \centering
        \renewcommand{\arraystretch}{1.5} 
        \begin{tabularx}{\textwidth}{|>{\arraybackslash}X|}
            \hline
            \textbf{Entity Labels} \\ \hline
            RESTAURANT, FOOD, DRINKS, AMBIENCE, SERVICE, LOCATION \\ \hline
            \textbf{Attribute Labels} \\ \hline
            GENERAL, PRICES, QUALITY, STYLE\_OPTIONS, MISCELLANEOUS
 \\ \hline
        \end{tabularx}
\end{subtable}
% \caption*{Restaurant Label Category}
\label{tab:Label Category}
\end{table}

%\vspace{-20pt}
%\clearpage
    \item Hotel
%\vspace{-8pt}
\begin{table}[H]
\small
\centering
\setlength{\tabcolsep}{6pt}
\renewcommand{\arraystretch}{1.0}
    \begin{subtable}{1.0\textwidth}
        \centering
        \renewcommand{\arraystretch}{1.5} 
        \begin{tabularx}{\textwidth}{|>{\arraybackslash}X|}
            \hline
            \textbf{Entity Labels} \\ \hline
            HOTEL, ROOMS, FACILITIES, ROOM\_AMENITIES, SERVICE, LOCATION, FOOD\_DRINKS \\ \hline
            \textbf{Attribute Labels} \\ \hline
            GENERAL, PRICE, COMFORT, CLEANLINESS, QUALITY, DESIGN\_FEATURES, STYLE\_OPTIONS, MISCELLANEOUS \\ \hline
        \end{tabularx}
        % \caption{Hotel}
    \end{subtable}
% \end{adjustbox}
% \caption*{Hotel}
\label{tab:Label Category}
\end{table}
\end{itemize}

\vspace{3em}

\section{Overview of Subtasks with Examples}
\label{sec:trackA-subtask-example}

\begin{table}[h!]
\centering
\small
\resizebox{1.0\linewidth}{!}{%
\begin{tabular}{lcccc}
\toprule
\textbf{Task} & \textbf{Input} & \textbf{Output} & \textbf{Prediction Type} & \textbf{Metric} \\
\midrule
\multirow{2}{*}[-0.6ex]{DimASR} & text + aspects& V\#A & \multirow{2}{*}{Regression} & \multirow{2}{*}[-0.6ex]{RMSE} \\\cmidrule(lr){2-2}\cmidrule(lr){3-3}
& The food was excellent & 8.00\#8.12 & & \\ \cmidrule(lr){1-5}

\multirow{2}{*}[-0.6ex]{DimASTE} & text & (A, O, V\#A) & Extraction & \multirow{2}{*}[-0.6ex]{cF1} \\\cmidrule(lr){2-2}\cmidrule(lr){3-3}
& Service at the bar was a little slow & Service, a little slow, 4.10\#4.30) & Regression& \\ \cmidrule(lr){1-5}

\multirow{3}{*}[-1.0ex]{DimASQP} & text & (A, C, O, V\#A) & Extraction & \multirow{3}{*}[-1.0ex]{cF1} \\\cmidrule(lr){2-2}\cmidrule(lr){3-3}
& \multirow{2}{*}[-0.6ex]{Their sodas are usually expired and flat} & (sodas, DRINKS\#QUALITY, usually expired, 1.90\#7.20)& Classification& \\
\addlinespace[1ex]
 & & (sodas, DRINKS\#QUALITY, flat, 2.40\#6.80) & Regression & \\

\bottomrule
\end{tabular}}
%\caption{\textbf{DimABSA subtasks} with input–output structure, task type, and metrics.}
\label{tab:dimabsa_task_summary}
\end{table}
\clearpage

\section{Example Calculation of cF1}
\label{sec:example_calculating_cF1}

\begin{table}[H]
\centering
\renewcommand{\arraystretch}{1.1}
\begin{tabular*}{\textwidth}{@{\extracolsep{\fill}}|l|c|c|c|c|}
\hline

\multirow{2}{*} &  & \multicolumn{2}{c|}{\textbf{VA error distance}} & \\ \cline{3-4}
 {\textbf{Prediction/Gold}} &\multicolumn{1}{p{1.5cm}|}{\centering $TP_{cat}$ \\ (A)} & \multicolumn{1}{p{1.5cm}|}{\centering Raw \\ (B)} & \multicolumn{1}{p{3cm}|}{\centering Normalized \\ $(C)=(B)/ \sqrt{128}$} & \multicolumn{1}{p{1.5cm}|}{\centering \textbf{cTP} \\ (A)-(C)}\\ \hline
 
P: (food, good, 8.00\#8.00) & \multirow{2}{*}{1} & \multirow{2}{*}{$\sqrt{2}$} & \multirow{2}{*}{$\frac{\sqrt{2}}{\sqrt{128}}= 0.125$} & \multirow{2}{*}{0.875} \\
G: (food, good, 7.00\#7.00) & & & & \\ \hline

P: (soup, spicy, 7.50\#7.50) & \multirow{2}{*}{1} & \multirow{2}{*}{$\sqrt{32}$} & \multirow{2}{*}{$\dfrac{\sqrt{32}}{\sqrt{128}} = 0.5$} & \multirow{2}{*}{0.5} \\
G: (soup, spicy, 3.50\#3.50) & & & & \\ \hline

P: (staff, friendly, 7.00\#7.00) & \multirow{2}{*}{0} & \multirow{2}{*}{-} & \multirow{2}{*}{-} & \multirow{2}{*}{0} \\
G: (staff, always friendly, 7.50\#7.50) & & & & \\ \hline

P: (staff, good, 7.00\#7.00) & \multirow{2}{*}{0} & \multirow{2}{*}{-} & \multirow{2}{*}{-} & \multirow{2}{*}{0} \\
G: N/A & & & & \\ \hline
\multicolumn{3}{|c|}{} & Total cTP& 1.375 \\ \hline
\multicolumn{5}{|r|}{$cRecall = 1.375/3 = 0.458$} \\ \hline
\multicolumn{5}{|r|}{$cPrecision = 1.375/4 = 0.344$}  \\ \hline
\multicolumn{5}{|r|}{$cF1= (2*0.458*0.344)/(0.458+0.344)=0.393$}  \\ \hline
\end{tabular*}
\end{table}
Note: The VA scores lie in the range [1, 9]. When the VA prediction is perfect (i.e., dist=0), cRecall/cPrecision reduces to the standard recall/precision. 

\clearpage

\section{Teams}
\centering
\small
\setlength{\tabcolsep}{4pt}
\renewcommand{\arraystretch}{1.15}

\begin{longtable*}{l p{0.15\textwidth} p{0.4\textwidth} l}
\toprule
Tracks & Team & Affiliation & Paper \\
\midrule
\endfirsthead

\multicolumn{4}{l}{\textit{(Continued from previous page)}} \\
\toprule
Tracks & Team & Affiliation & Paper \\
\midrule
\endhead

\midrule
\multicolumn{4}{r}{\textit{(Continued on next page)}} \\
\endfoot

\bottomrule
\endlastfoot

A & AILS-NTUA & Artificial Intelligence and Learning Systems Laboratory, School of Electrical and Computer Engineering, National Technical University of Athens, Greece & \cite{gazetas-semeval2026-sub180} \\
A & ALPS-Lab & Fujian University of Technology, China & \cite{dai-semeval2026-sub238} \\
A & Bert Kittens & Individual researcher & \cite{sukhodolsky-semeval2026-sub418} \\
B & CLRG & Bangladesh University of Engineering and Technology, Bangladesh; BRAC University, Bangladesh & \cite{iqbal-semeval2026-sub445} \\
B & CYUT & Chaoyang University of Technology, Taiwan & \cite{wu-semeval2026-sub26} \\
A, B & DUTH & Department of Electrical \& Computer Engineering, Democritus University of Thrace, Greece & \cite{arampatzis-semeval2026-sub94} \\
A, B & HUS@NLP-VNU & Hanoi University of Science, Vietnam; National University, Vietnam & \cite{cao-semeval2026-sub312} \\
A & Habib university & Habib University, Pakistan & \cite{affan-semeval2026-sub328} \\
A & ICT-NLP & Institute of Computing Technology, Chinese Academy of Sciences, China & \cite{huang-semeval2026-sub147} \\
A, B & LogSigma & University of Göttingen, Germany & \cite{hikal-semeval2026-sub182} \\
A & NCL-BU & Bournemouth University, UK; Newcastle University, UK & \cite{wu-semeval2026-sub270} \\
A, B & NTNU-SMIL & Speech and Machine Intelligence Laboratory (SMIL), Department of Computer Science and Information Engineering, National Taiwan Normal University, Taiwan & \cite{lin-semeval2026-sub3} \\
A & NYCU Speech Lab & Institute of Artificial Intelligence Innovation, National Yang Ming Chiao Tung University, Taiwan & \cite{hsieh-semeval2026-sub225} \\
A, B & PAI & Ping An Life Insurance Company of China, Ltd. & \cite{ruan-semeval2026-sub215} \\
A, B & PALI & none & \cite{chen-semeval2026-sub102} \\
A & PICT & Pune Institute of Computer Technology, India & \cite{bhalgat-semeval2026-sub50} \\
A & PUEB-DimASR & Poznan University of Economics and Business, Poland & \cite{riewepera-semeval2026-sub59} \\
A, B & Pixel Phantoms & Sri Sivasubramaniya Nadar College of Engineering, India; Loyola-ICAM College of Engineering and Technology, India & \cite{s-semeval2026-sub125} \\
A & QuadAI & Leiden University, Netherlands; Leiden University Medical Center (LUMC), Netherlands; Manchester Metropolitan University, UK & \cite{devink-semeval2026-sub11} \\
A & RPI Team & Rensselaer Polytechnic Institute, Troy NY, USA & \cite{modi-semeval2026-sub444} \\
A & SCUZANE & Sichuan University, China & \cite{zhou-semeval2026-sub284} \\
A, B & SCU\_Mesclab & Department of Data Science, Soochow University, Taiwan; Department of Computer Networks, Faculty of Electrical Engineering and Informatics, Technical University of Košice, Slovakia & \cite{lee-semeval2026-sub214} \\
A & SRCB & Ricoh Software Research Center (Beijing) Co., Ltd & \cite{li-semeval2026-sub48} \\
A, B & Scmhl5 & College of Computer Science and Software Engineering, Shenzhen University, China & \cite{chen-semeval2026-sub249} \\
A & SokraTUM & Technical University of Munich, Germany & \cite{laschenko-semeval2026-sub271} \\
A & Takoyaki & CyberAgent, Japan & \cite{yamada-semeval2026-sub246} \\
A & Team BOBW (Best Of Both Worlds) & IT University of Copenhagen, Denmark & \cite{rynowiecki-semeval2026-sub199} \\
A & Team HausaNLP & National Open University of Nigeria, Nigeria; Gombe State University, Nigeria; Nassarawa State University Keffi, Nigeria; Nile University Abuja, Nigeria & \cite{adam-semeval2026-sub38} \\
A & Team VYN & Department of Computer Science \& Engineering, University of Moratuwa, Sri Lanka & \cite{thenuwara-semeval2026-sub250} \\
A & TeamLasse & Technical University of Munich, Germany & \cite{strothe-semeval2026-sub306} \\
A & TeleAI & Institute of Artificial Intelligence (TeleAI), China Telecom & \cite{zhou-semeval2026-sub262} \\
A & The Classics & HSE University, Russia & \cite{alshawi-semeval2026-sub371} \\
A & UNF-BMI & University of North Florida, USA & \cite{jones-semeval2026-sub357} \\
A & YNU-ABSA & Yunnan University, China & \cite{he-semeval2026-sub167} \\
A, B & YangS\_team & Chunghwa Telecom Co., Ltd., Taiwan
 & \cite{yang-semeval2026-sub221} \\
A & hdharpure & Indian Institute of Technology Patna, India & \cite{dharpure-semeval2026-sub315} \\
B & hllwan & Nanjing University of Science and Technology, China & \cite{li-semeval2026-sub174} \\
A & kevinyu66 & National Cheng Kung University, Taiwan & \cite{yu-semeval2026-sub17} \\
A & kirito & Yunnan University, China & \cite{hu-semeval2026-sub140} \\
A & looploop & Yunnan University, China & \cite{yang-semeval2026-sub218} \\
A & nchellwig & Media Informatics Group, University of Regensburg, Germany & \cite{hellwig-semeval2026-sub6} \\
A & ssurface3 & Skoltech, Russia & \cite{frolov-semeval2026-sub403} \\

\end{longtable*}
\captionof{table}{Participants information (tasks, affiliations, and papers).}
\label{tab:dimabsa-participants-info}

\clearpage
\section{Leaderboards}
% Auto-generated by dimabsa_results_table_to_txt.py
% Requires: \usepackage{booktabs} and \usepackage{graphicx}

\begin{table*}[h]
\centering
\small
\setlength{\tabcolsep}{3pt}
\renewcommand{\arraystretch}{1.15}
\begin{tabular}{r l r r r r r r r r r r}
\toprule
S/N & Team & eng-rest & eng-lap & jpn-hot & jpn-fin & rus-rest & tat-rest & ukr-rest & zho-rest & zho-lap & zho-fin \\
\midrule
1 & AILS-NTUA & 1.3933 & 1.4401 & 0.7484 & 0.9635 & 1.7236 & 2.1144 & 1.6724 & 1.0023 & 0.7457 & 0.5425 \\
2 & Bert Kittens & 1.1812 & 1.2769 & 0.7267 & 0.9675 & 1.5828 & 2.2118 &  &  &  &  \\
3 & DUTH &  & 1.5924 &  &  &  &  &  &  &  &  \\
4 & HUS@NLP-VNU & 1.2745 & 1.4109 & 0.6386 & 0.8296 & 1.3075 & 1.8220 & 1.3538 & 0.9595 & 0.6663 & \textbf{0.4841} \\
5 & Habib university & 1.3049 & 1.3654 & 0.6680 & 0.8907 & 1.4344 & 1.6041 & 1.4661 & 0.9898 & 0.7311 & 0.5333 \\
8 & LogSigma & \textbf{1.1035} & \textbf{1.2408} &  &  &  &  &  &  &  &  \\
9 & NCL-BU & 1.4861 & 1.4562 &  &  &  &  &  & 0.9553 & 0.7510 & 0.5391 \\
11 & NTNU-SMIL & 1.2846 & 1.3501 & 0.6378 & 0.9278 & 1.4430 & 2.1785 & 1.4655 & 0.9841 & 0.6695 & 0.5115 \\
12 & PAI & 1.2141 & 1.4394 & 0.6508 & 0.7584 & \textbf{1.2190} & \textbf{1.5294} & \textbf{1.1888} & 0.9766 & 0.6800 & 0.5977 \\
13 & PALI & 1.2866 & 1.3612 & 0.6237 & 0.7532 & 1.3642 & 1.7121 & 1.4030 & 0.9805 & 0.6681 & 0.6042 \\
14 & PICT & 1.1958 & 1.3261 &  &  &  &  &  &  &  &  \\
15 & PUEB-DimASR & 1.7011 & 1.7587 & 1.2827 & 1.4505 & 2.2749 & 2.3347 & 2.2589 & 1.2405 & 1.1343 & 0.8179 \\
16 & Pixel Phantoms & 1.3656 & 1.4190 & 0.7297 & 1.0242 & 1.7686 & 2.0729 & 1.5937 & 0.9823 & 0.7438 & 0.7259 \\
17 & QuadAI & 1.3632 & 1.4062 &  &  &  &  &  &  &  &  \\
18 & RPI Team & 1.2006 & 1.2833 & 0.6413 & 0.8254 & 1.4849 & 1.7837 & 1.5485 & 0.9599 & 0.7005 & 0.5398 \\
19 & SCUZANE & 1.3483 & 1.4242 & 0.7129 & 0.9580 & 1.5572 & 2.3199 & 1.5730 & 0.9636 & 0.6981 & 0.5117 \\
20 & SCU\_Mesclab & 1.2277 & 1.3946 &  &  &  &  &  & 1.1210 & 0.9222 & 0.6692 \\
21 & SRCB & 1.2270 &  &  &  &  &  &  &  &  &  \\
22 & Scmhl5 & 1.3168 &  & 0.6811 & 0.9292 & 1.4609 & 2.0142 & 1.4732 & 0.9838 & 0.7165 &  \\
23 & SokraTUM & 1.3011 & 1.2942 &  &  &  &  &  &  &  &  \\
24 & Team HausaNLP & 1.4936 & 1.5143 &  &  &  &  &  &  &  &  \\
25 & Team VYN & 1.7978 &  &  &  &  &  &  &  &  &  \\
26 & TeamLasse & 1.4265 &  &  & 0.9982 & 1.5991 & 2.0212 & 1.6039 & 1.1601 & 1.0931 &  \\
27 & TeleAI & 1.2139 & 1.2425 & \textbf{0.5561} & \textbf{0.6581} & 1.2456 & 1.7662 & 1.3234 & 0.9265 & \textbf{0.6103} & 0.4866 \\
28 & The Classics & 1.2324 & 1.3283 &  &  & 1.6390 &  &  &  &  &  \\
29 & UNF-BMI & 1.3920 & 1.4336 &  &  &  &  &  &  &  &  \\
30 & YNU-ABSA & 1.4001 & 1.4198 & 0.7554 & 1.0026 & 1.5967 & 2.0104 &  & 0.9945 &  &  \\
31 & YangS\_team & 1.2772 & 1.3455 &  &  &  &  &  & 0.9433 & 0.6867 & 0.4864 \\
32 & hdharpure & 1.5003 & 1.5412 & 0.8378 & 1.0292 & 1.6515 & 2.0463 & 1.7172 & 0.9847 & 0.7902 & 0.5704 \\
33 & kirito & 1.3966 & 1.5010 &  &  &  &  &  &  &  &  \\
34 & looploop & 1.2048 & 1.3021 &  &  &  &  &  &  &  &  \\
36 & ssurface3 & 1.9115 & 1.8486 & 1.1509 & 1.4514 & 1.7572 & 1.9471 & 1.7793 & 1.0870 & 0.9482 & 0.8329 \\
\midrule
 & Average & 1.3508 & 1.4110 & 0.7408 & 0.9531 & 1.5471 & 1.9482 & 1.5467 & 1.0008 & 0.7628 & 0.5805 \\
\midrule
 & Baseline (Kimi-K2 Thinking) & 2.1461 & 2.1893 & 1.7553 & 1.6396 & 1.7768 & 1.9380 & 1.7805 & 1.8959 & 1.6440 & 1.9652 \\
 & Baseline (Qwen-3 14B) & 2.6427 & 2.8089 & 2.2906 & 1.8964 & 2.1528 & 2.6367 & 2.2121 & 2.0073 & 1.7706 & 1.4707 \\
\bottomrule
\end{tabular}
\caption{DimABSA results (Track A, Subtask 1).}
\label{tab:dimabsa-track-a-subtask-1}
\end{table*}

\begin{table*}[h]
\centering
\small
\setlength{\tabcolsep}{3pt}
\renewcommand{\arraystretch}{1.15}
\begin{tabular}{r l r r r r r r r r}
\toprule
S/N & Team & eng-rest & eng-lap & jpn-hot & rus-rest & tat-rest & ukr-rest & zho-rest & zho-lap \\
\midrule
1 & AILS-NTUA & 0.6518 & 0.5311 & 0.5021 & 0.4988 & 0.3874 & 0.4725 & 0.5042 & 0.4646 \\
2 & ALPS-Lab & 0.0000 & 0.0000 & 0.0000 & 0.5414 & 0.4798 & 0.5613 & 0.5247 & 0.4935 \\
3 & Bert Kittens & 0.5628 & 0.4469 & 0.4202 & 0.3137 & 0.1692 &  &  &  \\
5 & HUS@NLP-VNU & 0.6391 & 0.5304 &  &  &  &  &  &  \\
6 & Habib university & 0.5202 & 0.4770 & 0.3311 & 0.5492 & 0.4839 & 0.5324 & 0.4622 & 0.4159 \\
7 & ICT-NLP & 0.6174 & 0.5622 & 0.3152 & 0.4622 & 0.3088 & 0.4355 & 0.2756 & 0.3019 \\
9 & PAI & 0.6903 & 0.6169 & 0.5682 & \textbf{0.5793} & 0.4908 & \textbf{0.5787} & \textbf{0.5638} & 0.5306 \\
10 & PALI & 0.6928 & 0.6242 & 0.5666 & 0.5724 & 0.4828 & 0.5671 & 0.5634 & \textbf{0.5308} \\
11 & Pixel Phantoms & 0.0265 &  &  &  &  &  &  &  \\
12 & Scmhl5 & 0.6127 & 0.5136 & 0.3357 & 0.3960 & 0.3649 & 0.4267 & 0.3955 & 0.3111 \\
13 & SokraTUM & 0.6326 & 0.5635 &  &  &  &  &  &  \\
14 & Takoyaki & \textbf{0.7021} & \textbf{0.6366} & 0.5340 & 0.5564 & 0.5092 & 0.5438 & 0.5382 & 0.4758 \\
15 & TeamLasse & 0.6391 & 0.5513 & 0.5694 & 0.5253 & 0.4496 & 0.5270 & 0.5320 & 0.4807 \\
16 & TeleAI & 0.6294 & 0.5345 & \textbf{0.5837} & 0.5736 & 0.4863 & 0.5712 & 0.5448 & 0.5292 \\
17 & The Classics & 0.5650 & 0.4763 &  &  &  &  &  &  \\
18 & YNU-ABSA & 0.5240 & 0.4952 &  &  &  &  &  &  \\
19 & kevinyu66 & 0.6707 & 0.5503 & 0.5366 & 0.5117 & 0.3731 & 0.4865 & 0.5089 & 0.4802 \\
20 & kirito & 0.5676 & 0.4733 &  &  &  &  &  &  \\
21 & looploop & 0.5799 & 0.4799 &  &  &  &  &  &  \\
22 & nchellwig & 0.6985 & 0.6092 & 0.5518 & 0.5640 & \textbf{0.5119} & 0.5285 & 0.5488 & 0.5110 \\
\midrule
 & Average & 0.5686 & 0.5123 & 0.4570 & 0.5106 & 0.4223 & 0.5183 & 0.4973 & 0.4606 \\
\midrule
 & Baseline (Kimi-K2 Thinking) & 0.4920 & 0.4424 & 0.3464 & 0.4242 & 0.3577 & 0.4220 & 0.3529 & 0.2494 \\
 & Baseline (Qwen-3 14B) & 0.4483 & 0.3827 & 0.1622 & 0.3341 & 0.2020 & 0.3099 & 0.2509 & 0.2099 \\
\bottomrule
\end{tabular}
\caption{DimABSA results (Track A, Subtask 2).}
\label{tab:dimabsa-track-a-subtask-2}
\end{table*}

\begin{table*}[h]
\centering
\small
\setlength{\tabcolsep}{3pt}
\renewcommand{\arraystretch}{1.15}
\begin{tabular}{r l r r r r r r r r}
\toprule
S/N & Team & eng-rest & eng-lap & jpn-hot & rus-rest & tat-rest & ukr-rest & zho-rest & zho-lap \\
\midrule
1 & AILS-NTUA & 0.5988 & 0.2694 & 0.3747 & 0.4369 & 0.3306 & 0.4154 & 0.4544 & 0.3703 \\
2 & ALPS-Lab & 0.6202 & 0.3395 & 0.3617 & 0.5042 & 0.4404 & 0.5163 & 0.4853 & 0.3968 \\
3 & Bert Kittens & 0.5162 & 0.2578 & 0.2845 &  & 0.1479 &  &  &  \\
4 & HUS@NLP-VNU & 0.5871 & 0.2587 &  &  &  &  &  &  \\
5 & Habib university & 0.0000 & 0.0000 & 0.1853 & 0.3029 & 0.2500 & 0.2938 & 0.4199 & 0.3139 \\
7 & NYCU Speech Lab &  &  &  &  &  &  & \textbf{0.5521} & \textbf{0.4824} \\
8 & PAI &  & 0.3758 &  & \textbf{0.5599} & 0.4523 & \textbf{0.5437} & 0.5360 & 0.4316 \\
9 & PALI & 0.6395 & 0.3793 & \textbf{0.4252} & 0.5496 & 0.4443 & 0.5307 & 0.5357 & 0.4319 \\
10 & Scmhl5 & 0.5119 & 0.2752 & 0.2195 & 0.3138 & 0.2629 & 0.3384 & 0.3309 & 0.1996 \\
11 & SokraTUM & 0.5612 & 0.2512 &  &  &  &  &  &  \\
12 & Takoyaki & \textbf{0.6514} & \textbf{0.4227} & 0.4086 & 0.5130 & \textbf{0.4736} & 0.5019 & 0.4966 & 0.3745 \\
13 & Team BOBW & 0.5317 & 0.2317 &  &  &  &  &  &  \\
14 & TeamLasse & 0.5937 & 0.3049 & 0.3992 & 0.4991 & 0.4113 & 0.4879 & 0.5026 & 0.3478 \\
15 & TeleAI & 0.5487 & 0.3281 & 0.1258 & 0.3357 & 0.2512 & 0.3245 & 0.3979 & 0.1885 \\
16 & The Classics &  & 0.3072 &  &  &  &  &  &  \\
17 & YNU-ABSA & 0.5183 &  &  &  &  &  &  &  \\
18 & kirito & 0.5201 & 0.2480 &  &  &  &  &  &  \\
19 & looploop & 0.5562 & 0.2781 &  &  &  &  &  &  \\
20 & nchellwig & 0.6403 & 0.4006 & 0.3974 & 0.5083 & 0.4557 & 0.4746 & 0.4966 & 0.4016 \\
\midrule
 & Average & 0.5398 & 0.2908 & 0.3259 & 0.4526 & 0.3581 & 0.4446 & 0.4728 & 0.3602 \\
\midrule
 & Baseline (Kimi-K2 Thinking) & 0.3746 & 0.2795 & 0.1943 & 0.2963 & 0.2380 & 0.2971 & 0.2859 & 0.1900 \\
 & Baseline (Qwen-3 14B) & 0.2673 & 0.1529 & 0.0400 & 0.1682 & 0.0954 & 0.1641 & 0.1605 & 0.1124 \\
\bottomrule
\end{tabular}
\caption{DimABSA results (Track A, Subtask 3).}
\label{tab:dimabsa-track-a-subtask-3}
\end{table*}
\clearpage
\begin{table*}[t]
\centering
\small
\setlength{\tabcolsep}{3pt}
\renewcommand{\arraystretch}{1.15}
\begin{tabular}{r l r r r r r}
\toprule
S/N & Team & eng-env & deu-pol & zho-env & pcm-pol & swa-pol \\
\midrule
1 & CLRG & 2.0654 & 2.0923 & 0.6170 & 1.9114 & 2.1320 \\
2 & CYUT & 1.6331 & 1.4827 & 0.6771 & \textbf{1.1024} & 2.1042 \\
3 & DUTH & 2.1964 &  &  &  &  \\
4 & HUS@NLP-VNU & 1.6899 & 1.4108 & 0.5826 & 1.4269 & 1.8713 \\
5 & LogSigma & \textbf{1.4734} & \textbf{1.3417} & 0.6460 & 1.1269 & \textbf{1.7959} \\
6 & NTNU-SMIL & 1.5207 & 1.3467 & 0.5561 & 1.5674 & 1.9602 \\
7 & PAI & 1.6768 & 1.5110 & 0.6269 & 1.1399 & 2.2519 \\
8 & PALI & 1.8048 & 1.5688 & 0.7047 & 1.4078 & 2.4544 \\
9 & Pixel Phantoms & 2.0893 & 1.5509 & 0.7364 & 1.7878 & 2.2700 \\
10 & SCU\_Mesclab & 1.5714 &  & 0.7452 &  &  \\
11 & Scmhl5 & 1.6612 & 1.4375 & 0.6765 & 1.4072 & 1.9391 \\
12 & YangS\_team & 1.5731 &  & \textbf{0.5468} &  &  \\
13 & hllwan & 1.5122 & 1.4937 & 0.6154 & 1.2232 & 1.9522 \\
\midrule
 & Average & 1.7283 & 1.5486 & 0.6401 & 1.4551 & 2.0731 \\
\midrule
 & Baseline (Mistral-3 14B) & 1.6430 & 1.5910 & 0.7400 & 1.7390 & 2.2990 \\
 & Baseline (mBERT) & 2.6985 & 2.3294 & 1.2756 & 3.2152 & 2.7835 \\
\bottomrule
\end{tabular}
\caption{DimStance results (Track B).}
\label{tab:dimabsa-track-b-subtask-1}
\end{table*}
\clearpage

\section{System Statistics}

\vspace{1em}

\begin{figure}[h]
    \centering
    \includegraphics[height=0.35\textheight,keepaspectratio]{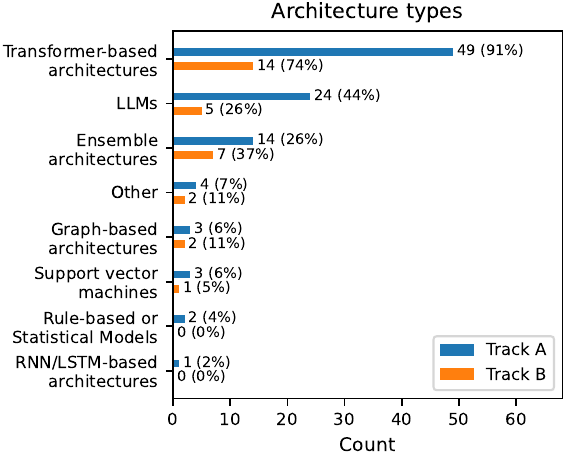}
    \caption{\textbf{Model architectures used by participants.}}
        \label{fig:architectures}
\end{figure}

\vspace{6em}

\begin{figure}[!htbp]
    \centering
    \includegraphics[height=0.35\textheight,keepaspectratio]{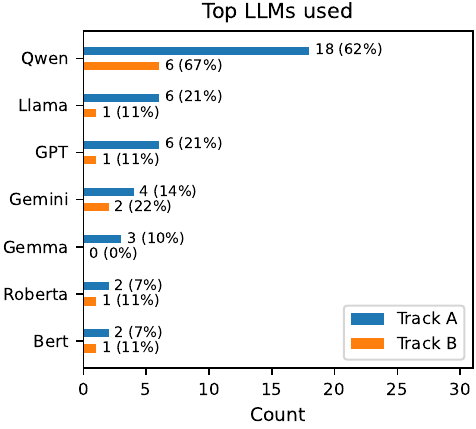}
    \caption{\textbf{LLMs used by participants.}}
        \label{fig:models}
\end{figure}

\begin{figure}[h]
    \centering
    \includegraphics[height=0.35\textheight,keepaspectratio]{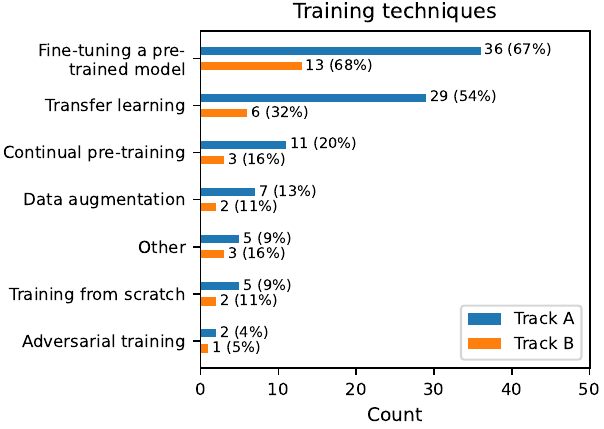}
    \caption{\textbf{Training techniques used by participants.}}
        \label{fig:trainings}
\end{figure}

\begin{figure}[h]
    \centering
    \includegraphics[height=0.35\textheight,keepaspectratio]{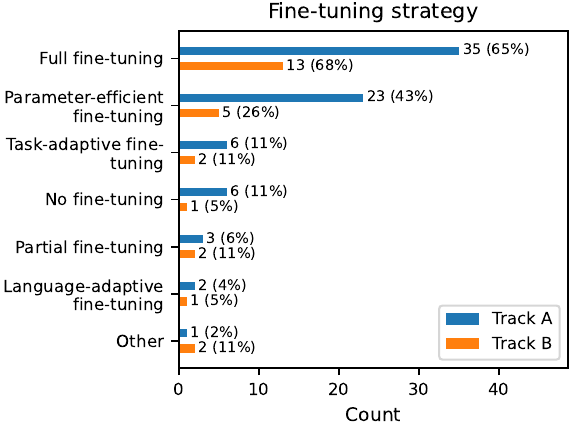}
    \caption{\textbf{Fine-tuning strategies used by participants.}}
        \label{fig:finetuning}
\end{figure}

\begin{figure}[h]
    \centering
    \includegraphics[height=0.35\textheight,keepaspectratio]{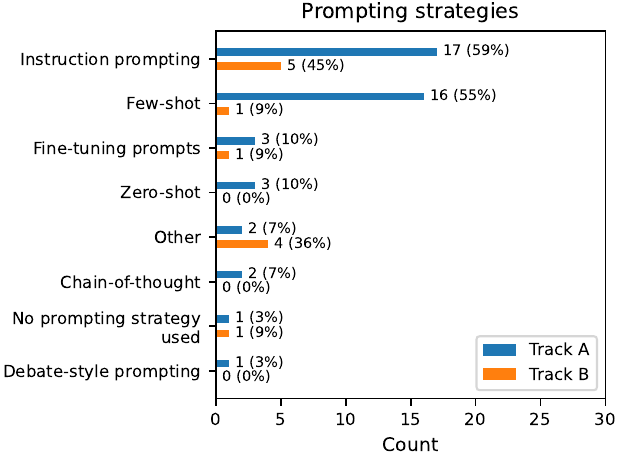}
    \caption{\textbf{Prompting strategies used by participants.}}
        \label{fig:prompting}
\end{figure}

\begin{figure}[h]
    \centering
    \includegraphics[height=0.26\textheight,keepaspectratio]{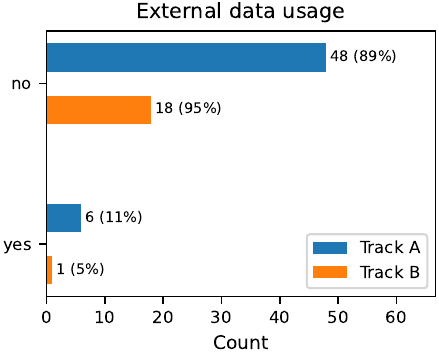}
    \caption{\textbf{External data used by participants.}}
        \label{fig:external_data}
\end{figure}

\end{document}